\documentclass[10pt,journal,compsoc]{IEEEtran}
\pdfoutput=1

\ifCLASSOPTIONcompsoc
  \usepackage[nocompress]{cite}
\else
  \usepackage{cite}
\fi

\usepackage{bm}     
\usepackage{dsfont} 

\usepackage{color} 

\usepackage{amssymb}
\usepackage{amsmath}
\usepackage[ruled,vlined]{algorithm2e}
\makeatletter
\newcommand{\removelatexerror}{\let\@latex@error\@gobble}
\makeatother

\usepackage{multirow}

\DeclareMathOperator{\poly}{poly}
\DeclareMathOperator*{\argmin}{argmin}

\ifCLASSOPTIONcompsoc
\usepackage[caption=false,font=normalsize,labelfont=sf,textfont=sf]{subfig}
\else
\usepackage[caption=false,font=footnotesize]{subfig}
\fi
\usepackage{graphicx}

\def\authorslong{Thomas~Lang, Tomas~Sauer}

\def\titlename{Geometric Active Learning for\\ Segmentation of Large 3D Volumes}

\usepackage{filecontents}

\newtheorem{definition}{Definition}
\newtheorem{lemma}{Lemma}

\begin{document}
\title{\titlename{}}
\author{\authorslong{}%
\IEEEcompsocitemizethanks{\IEEEcompsocthanksitem T. Lang is with 
  FORWISS, University of Passau, and the Fraunhofer
  Institute of Integrated Circuits, Division Development Center X-ray
  Technology.\protect\\
  E-mail: langthom@forwiss.uni-passau.de%
  \IEEEcompsocthanksitem T. Sauer is with 
  the Chair of Digital Image Processing \& FORWISS, University of
  Passau, and the Fraunhofer Institute of Integrated Circuits, Division
  Development Center X-ray Technology.\protect\\ 
  E-mail: tomas.sauer@uni-passau.de}%
\thanks{Manuscript received ??; revised ??.}}

\markboth{}{Lang \MakeLowercase{\textit{et al.}}: \titlename{}}

\IEEEtitleabstractindextext{%
\begin{abstract}
  Segmentation, i.e., the partitioning of volumetric data into
  components, is a crucial task in many image processing applications
  ever since such data could be generated.
  Most existing applications nowadays, specifically CNNs, make use of voxelwise
  classification systems which need to 
  be trained on a large number of annotated training volumes.
  However, in many practical applications such data sets are seldom available
  and the generation of annotations is time-consuming and cumbersome.
  In this paper, we introduce a novel voxelwise segmentation method
  based on active learning on geometric features.
  Our method uses interactively provided seed points to train a voxelwise 
  classifier based entirely on local information. The combination of
  an ad hoc incorporation of domain knowledge and local processing
  results in a flexible yet efficient segmentation method that is applicable
  to three-dimensional volumes without size restrictions. We illustrate
  the potential and flexibility of our approach by applying it to
  selected computed tomography scans where we perform different
  segmentation tasks to scans from different domains and of different
  sizes.
\end{abstract}

}

\maketitle
\IEEEdisplaynontitleabstractindextext
\IEEEpeerreviewmaketitle

\ifCLASSOPTIONcompsoc
\IEEEraisesectionheading{\section{Introduction}\label{sec:introduction}}
\else
\section{Introduction}
\label{sec:introduction}
\fi

\IEEEPARstart{E}{ver} since the invention of computed tomography (CT)
and the resulting generation of three-dimensional data, the extraction
of certain components of such an image has been the core problem in
different applications, among (many) others the detection of malign
structures or visual inspection of selected parts.
In clinical tomography, segmentation is a common task for which often
large datasets, including annotated ground truth data, exist. In
contrast, industrial applications typically not only generate much
larger volumes at higher resolutions, at the same time annotated data
is rarely available. Moreover, even the ``certain'' is uncertain in
the sense that specifying what exactly has to be segmented from a
volume is difficult in many applications. In other words, finding the
right mode of user interaction also becomes an issue when dealing with
industrial CT data.

This inhibits the usage of many modern segmentation methods
including atlas-based methods or neural networks and their variants,
since they, typically, require a large number of training samples
including ground truth in order to learn properly.
Another class of methods, including graph-based models and active
contour methods, are of a \emph{global} nature and thus are not
appropriate for segmenting big volumes whose size can easily exceed
the avaiable main memory.
As a result, many applications are forced to rely on two-dimensional
slice-wise image processing only, which easily overlooks thin
three-dimensional structures.
In addition, slice-based methods typically depend on the orientation
of the object in the volume or of the object during the scan.
Therefore, three-dimensional methods
are preferable which in turn enforces us to perform local processing
in order to handle large 3D volumes.

This paper introduces an active learning approach for interactive
segmentation of three-dimensional volumes without size restrictions,
hence constitutes an effort to overcome some limitations of related
approaches.
  To minimize the user interaction and the amount of data needed for
  training, we decided to use a Support Vector Machine (SVM) with
  features that are designed to capture local structures.
It is organized as follows.
In Section~\ref{sec:method}, we first revisit feature design with an
explicit emphasis on geometry. Next, we briefly recapitulate Support
Vector Machines, after which we introduce our segmentation
procedure. We will further give a coarse analysis of the algorithm's
asymptotic runtime and memory behavior 
to show that our method is well suited for the segmentation of large
volumes. 
In addition, we will also introduce a multiresolution segmentation
scheme based on an iteration of our segmentation technique, which
combines the advantage of faster volume processing with the
flexibility of our SVM segmentation algorithm. We will intuitively 
describe that the multiscale procedure is also applicable to volumes
without size restrictions. 
Section~\ref{sec:results} shows the performance our scheme on selected
datasets of different domains and sizes, illustrating both the
segmentation performance as well as qualitative results. 
We will conclude our contributions in Section~\ref{sec:conclusion}.

\section{Related Work}
Active learning (AL) is an emerging branch of machine learning and
particularly well suited for the segmentation of unknown objects where
one lacks annotated training data.
Existing work in the field of deep learning includes the 3D U-Net
architecture~\cite{UNet} which features a neural network to be trained
from few annotated slices while still obtaining a dense three-dimensional
segmentation result. 
More focussed on two-dimensional images than on volumes, the work
of~\cite{DLAL} uses a convolutional neural network approach that
includes a probability map prediction from which a pixelwise
uncertainty map is computed that serves as main indicator which yet
unlabeled images should be interactively labeled next. 
In a similar way,~\cite{ViewAL} proposed a viewpoint learning approach
in 2D images where view entropy and view divergence serve as
indicators which superpixel shall be selected for labeling. 
From a medical perspective, \cite{GANAL} focusses on lung nodule detection and
trains a conditional Generative Adversarial Network where the
generator creates realistic X-ray images from which a Bayesian neural
network computes its information content. The best images with respect
to that metric are then added to the training dataset.
Besides the popular neural network approaches, \cite{SVMAL1}
proposed an active learning procedure for text
processing, in which an optimal Support Vector Machine configuration
is determined by actively labeling instances closest to the separating
hyperplane. This method roughly corresponds to the popular uncertainty
sampling which is widely used in active
learning~\cite{ActiveLearningSurvey}.  
Similar approaches are used by~\cite{SVMAL2} and~\cite{SVMAL3}, while~\cite{SVMAL4}
uses the same idea by selecting, in each iteration, a
fixed size pool of unlabeled instances whose uncertainty is highest.
While~\cite{SVMAL2} describes a general way of how to employ active
learning in combination with Support Vector Machines, \cite{SVMAL3}
applies this to the field of image retrieval by using active learning to bootstrap
the classifier. In contrast to that, \cite{SVMAL4} requires
interactively labeled samples in order to adapt a classifier to a new
domain. 
Another similar approach was proposed in~\cite{SVMAL5} where an interactive
SVM approach is used to delineate object boundaries.
Returning to the field of image processing,~\cite{SVMInteractive2D}
aims to segment an image pixelwise into multiple classes based on
superpixel features. There, interactive user input in form of brush
strokes serves as seed selection from which color features are
extracted. A classifier is trained on these features and the given
superpixels are classified with it. 
Switching to the three-dimensional case,~\cite{SVMInteractive3D} uses
interactively provided markers as input for constructing a graph cut
segmentation problem which upon solving partitions the dataset into the
different categories of tumor, liver and other objects.
One of the approaches closest to our method is~\cite{GeomAL} which
incorporates geometric uncertainties into an active learning procedure
in order to classify supervoxels. However, their work does not
actually consider similarities to geometric primitives but rather
constructs some uncertainty measure based on random walks to select
the next supervoxels that shall be labeled. 

The related approaches we mention above are still not applicable to
our use case. The reason is that some of them require a full corpus of
annotated training datasets, specifically the methods incorporating
neural networks or atlas-based methods. 
In our situation, on the other hand, we often do not have any training
data at all as many of the scans are unique. Approaches which aim for
optimal segmentation by, e.g., using graph cut methods are not
feasible as well for processing large volumes which may no longer fit
into main memory. In this situation, constructing an even larger graph
from the volume and searching for a globally optimal solution is
unlikely to be processable. 
Instead, we focus on purely local processing and requesting as few
labeling effort as possible. 
To do so, we increase the discriminative power of our classifier by
explicitly using similarity of local regions to geometric primitives,
which to our knowledge was not used elsewhere for segmentation
purposes.

\section{Method}\label{sec:method}
Now we introduce our approach which combines both
machine learning techniques and classical methods and is based on an
active learning technique that
incorporates explicit features by considering local geometry.

\subsection{Geometric Features}
Our classification system relies on features obtained from \emph{local
  environments}. Precisely, around each voxel in the source volume, a
small voxel environment of configurable size $K\times K\times K$, for
$K > 1$ and odd, is used for the computation of features.
The choice of $K$ is a trade-off between locality and detectability of certain
features and it needs to be chosen dependent on the \emph{resolution},
not the \emph{size} of the scan. As a rule of thumb, $K$ should be
large enough to distinguish between the basic features, like
planarity, on a local basis, i.e., $K$ should be larger than the
thickness of what is considered to be an essentially two-dimensional object.
If no full environment can be extracted, the voxel is not processed
and the grayscale value at the according position in the target volume
is set to zero.
This is in accordance with our internal framework
constraints but any other extension like padding can be incorporated
as well with the usual price of boundary artifacts.
In the sequel, we will briefly explain all features that we use, but
we want to emphasize that not all features might
be used for every scan. In practice, it turned out that a domain
expert's proper choice of a subset of features best-suited to describe the
selected regions can significantly accelerate the segmentation
procedure. Alternatively, an automatic selection of relevant features
is possible by linear standard methods like the PCA,
see~\cite{FeatureSelectionPCA}. 

\medskip
Roughly, we distinguish between structural and geometric features in
our approach.
\emph{Structural features} capture properties of the grayscale
distribution of the local environment by its first four standardized
statistical moments~\cite{OnePassStatisticalMoments}. Without a doubt,
these are the features used most for grayscale image segmentation as
they express the mean voxel value, the standard deviation of these
values as well as the skewness and the (excess) kurtosis of the
distribution of voxel values in the local region. 

Depending on the concrete segmentation task, we may include the
position of the current voxel as a somewhat trivial feature. But this
can turn out to be useful if we try to separate objects touching each
other or segment objects heavily distorted by artifacts, for example.

In addition, we aim to encode the structure of the local region using
the well-known uniform local binary pattern for three orthogonal
planes (LBP/TOP)~\cite{LBPTOP,LBPMotiv}. In this feature, one
typically iterates over all pixels in a 2D region and computes the
signature (either one or zero) 
of the difference of the current pixel value against the center of the
region. The resulting bit string is converted into a single
integer. The uniform LBP version improves upon this by including that
in piecewise homogenous regions which are common in industrial
applications these bit strings are often of certain shapes
only. The improvement considers only the changes between patterns
which achieves some form of rotation invariance,
cf.~\cite{LBPUniform}. Since the LBP is classically defined for
two-dimensional images only, we slightly altered the
definition. First, we compute three projection images which are the
sums along one axis at a time. For each projection image, an LBP
descriptor is computed for each pixel in that image except at the
boundary. These descriptors are aggregated in a histogram. The
concatenation of the resulting three histograms form our structural
feature.

\medskip

Our main emphasis, however, lies on explicitly considering local
geometry in order to provide a flexible way of learning shape properties.
In particular, we include curvature as introduced in~\cite[Part
IV.A]{CurvatureWithoutIso} by iterating over each voxel inside the
local environment and computing the curvature inside a smaller
environment of dimensions $k\times k\times k$ around it, where the
choice $k=3$ performed well in our tests. The approach from
\cite{CurvatureWithoutIso} yields a  voxel-based curvature estimation
based on the second fundamental form without having 
to provide a threshold for the extraction of the surface. 
These curvature estimations are then aggregated in a histogram, thus
effectively tracking the distribution of curvatures.

As a contrast to curvature, we also consider planar or linear
structures which are particularly prominent in industrial
applications. There, we consider the $M$ nonzero voxels, more
precisely, voxels above a certain threshold, inside the local environment and 
fit either a line or a plane to it by least squares regression. Having
computed these structures, we collect the 
orthogonal distances of the nonzero voxels to the plane or line
structures into a histogram of distances.
Additionally, characteristics to represent planarity and width or
linearity and diameter, respectively, are computed in the following way.
We arrange the coordinates of these voxels columnwise in a matrix
$\bm{C}\in\mathbb{N}_0^{3\times M}$,  from which we compute the
centered matrix $\bm{C}' = \bm{C} - M^{-1}\bm{C11}^T$, 
where $\bm{1}$ is the vector of ones.
The three singular values of the matrix $\bm{C}'$ yield a simple but
efficient orientation feature. With a singular value decomposition
$\bm{C}' = \bm{U\Sigma V^T}$, we specifically need the column vector
$\bm{u}_1$ which points into the direction of the least squares
fitting line, and $\bm{u}_3$ which is the normal of the least squares
fitting plane. For both structures we then use the current voxel 
as a point lying on it. Denote by $L$ and $P$ be these structures for
linearity and planarity, respectively, and by
$D_L,D_P$ again the collections of orthogonal distances of the nonzero
voxels to the structures~\cite{OrthogonalDistance} which we arrange in
histograms $\mathcal{H}_L$ and $\mathcal{H}_P$,
respectively. Moreover, we compute the characteristics 
\begin{align*}
 f_1^p &= \exp\;(-\mathbb{E}[D_p]) \\
 f_2^p &= (2\sqrt{3}\lfloor K/2 \rfloor)^{-1}\;\left(\max_{d\in D_p} d
         - \min_{d\in D_p} d + 1\right) 
\end{align*}
for $p\in\{L,P\}$, where again $K$ denotes the size of the local
environment. The characteristic $f_1^p$ tends to 
one for local linear or planar structures, $p = L$ or $p=P$, and to zero otherwise,
hence it gives a measure for the linearity/planarity of the voxel region.
The width of the structure is indicated by $f_2^p$, whose denominator
represents the maximum possible distance 
of a nonzero voxel in the local environment to a structure passing
through the environment's center point. Eventually, the concatenation
$(f_1^p, f_2^p, \mathcal{H}_p), p\in\{L,P\}$ forms our \emph{linearity} and
\emph{planarity} features, respectively. 

\begin{figure*}[!t]
 \centering
 \subfloat[]{\includegraphics[width=.3\textwidth]{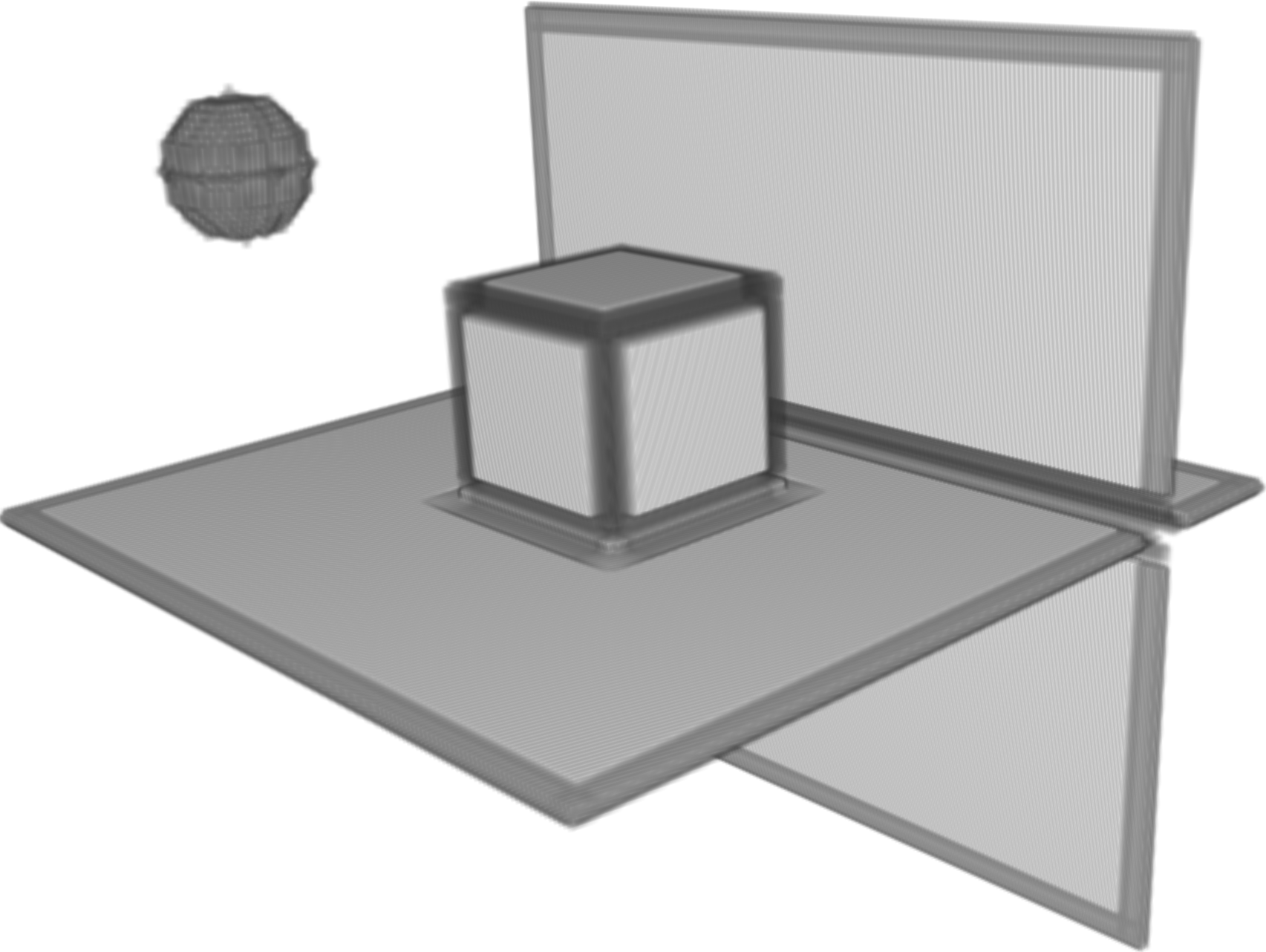}\label{fig:planarity}}
 \hfil
 \subfloat[]{\includegraphics[width=.3\textwidth]{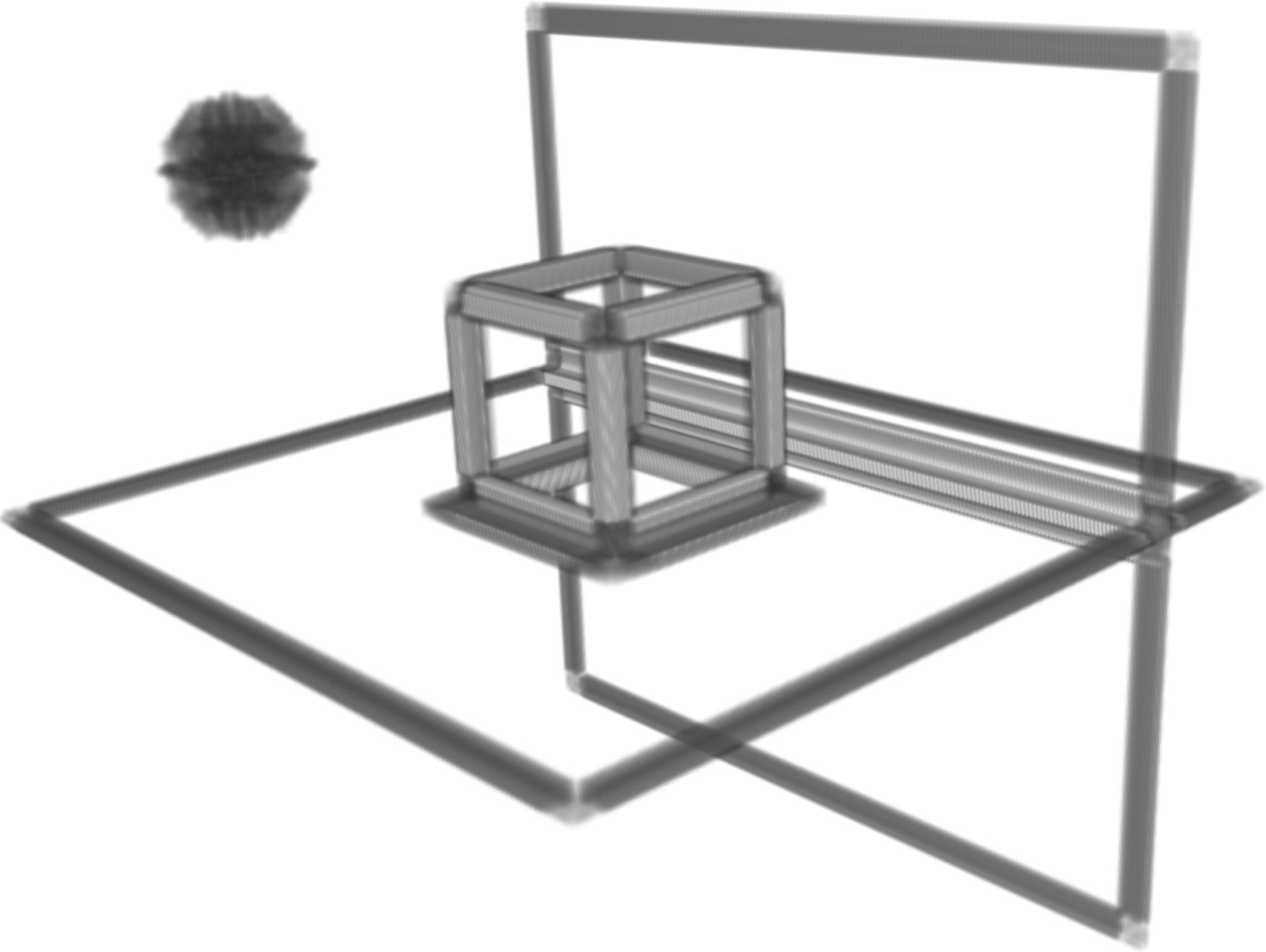}\label{fig:linearity}}
 \hfil
 \subfloat[]{\includegraphics[width=.3\textwidth]{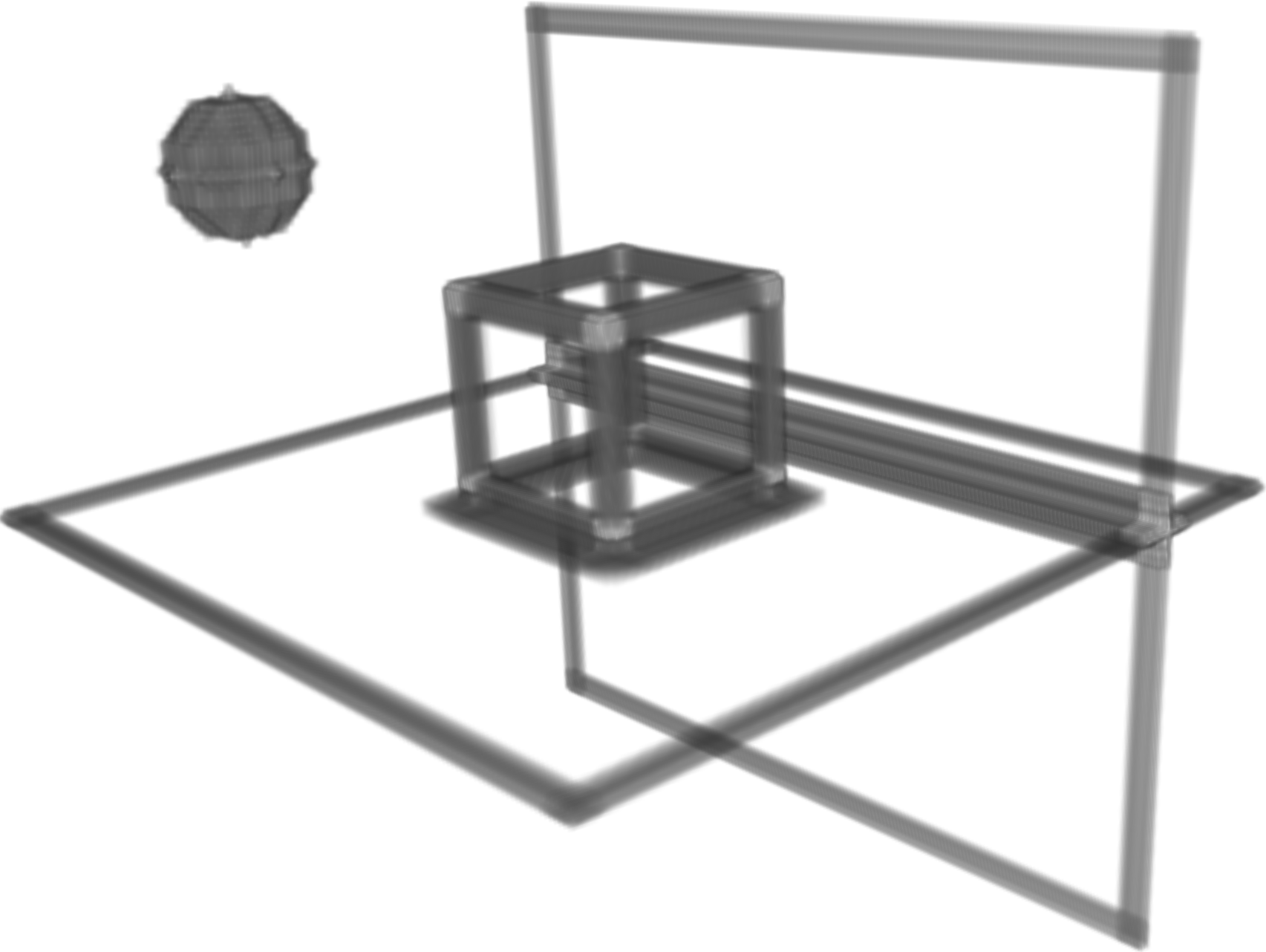}\label{fig:isotropy}}
 \caption{Inertia tensor features detecting geometry on a synthetic dataset;
  brighter colors indicate higher values.~\protect\subref{fig:planarity}
  shows the planarity $f_p$ which is high on the plane and also on the
  sides of the cube,~\protect\subref{fig:linearity} 
  depicts the linearity $f_l$ which enhances edges and~\protect\subref{fig:isotropy} 
  using $f_s$ detects isotropic regions.}
 \label{fig:inertiafeatures}
\end{figure*}

An alternative representation of planarity, linearity and also isotropy, 
is expressed via a variant of the inertia tensor as introduced
in~\cite[Eq. 8.11]{OrientationFeatures}. This representation computes
the Fourier tensor features from~\cite{LocalFourierFeatures} efficiently
in the spatial domain. A $3\times 3$ inertia tensor is built from the
partial derivatives giving a real symmetric matrix with
three nonnegative eigenvalues $\lambda_1 \geq \lambda_2 \geq \lambda_3 \geq 0$,
from which the geometric properties are computed as
\begin{displaymath}
 f_l = \frac{\lambda_1 - \lambda_2}{\lambda_1}, \quad
 f_p = \frac{\lambda_2 - \lambda_3}{\lambda_1}, \quad 
 f_s = \frac{\lambda_3}{\lambda_1}.
\end{displaymath}
The ability to detect geometry from this feature is illustrated in
Figure~\ref{fig:inertiafeatures}, where each voxel in a synthetic
dataset is assigned the according feature value at the respective position.
The \emph{linearity} $f_l$ is large if $\lambda_1 > \lambda_2 \approx \lambda_3 \approx 0$, i.e., when one direction
of inertia is very prominent while the other dimensions are close to
zero. A second dimension becomes prominent
in the presence of planar structures, and we have $\lambda_1 \approx
\lambda_2 > \lambda_3$ so that $f_p$ approaches $1$. Finally,
homogenous structures are identified by the lack of a dominant
direction, i.e., if all eigenvalues 
are approximately equal, in which case the \emph{isotropy} $f_s$
increases. Also note that the three nonnegative features add up to one
and thus are normalized relative to each other in a natural way.

Yet another type of histogram feature we use, is a simple histogram of
distances of nonzero voxels to the the center of the local environment
as an indication of the overall geometry. We chose this as an
alternative to the classical histogram of pairwise distances since the
latter has a quadratic runtime with respect to the local environment
size while the used feature is linear in it and thus easier to
compute, in particular as the size of the local environment is $K^3$ already.

To incorporate orientation we rely on the classical Histogram of
Oriented Gradients (HOG) as a feature. In its three-dimensional form,
one considers the estimated gradient vector and its azimuthal and
elevation angles which form a two-dimensional histogram where the
according bin is incremented by the Euclidean norm of any gradient
pointing in that direction. All orientations actually form a 3D
sphere, but an equidistantly partitioned planar 
two-dimensional histogram would have bins at the ``polar'' regions
where only very little orientations are covered. 
To compensate for this, we create a sphere histogram in which all
cells have the same surface area. In the concrete application, we
partitioned the elevation angles into five parts. 
The north-most and south-most sectors are further divided into four
partitions while the three middle sectors are divided into 14 bins. 
Next, for each voxel in the local environment, we estimate the
gradient at that point and compute the orientation angles from it. The
bin identified by these angles is then incremented by the gradient
magnitude provided that the latter
is larger than some small threshold to reduce the influence of noise.
The linearized sphere histogram then forms our HOG descriptor.

\medskip
Regarding the usage of features, we want to point out two small but important
caveats one has to take care of. 

Firstly, scaling the features is very important when using a feature
vector in a distance-based classifier, since one feature with a large
variance can render other features useless although they are important
geometry detectors. We recommend standardization of the features using
the well-established z-score~\cite{zscore} technique on all features
that belong together conceptually. Specifically, feature vector
elements which stem from a histogram should be scaled using the same
mean and standard deviation, while single values like the mean
grayscale value, should just be scaled using the according parameters
for this value.

Another more serious point is the comparison of histograms. It is
well-known that histograms can and should be interpreted as 
discretized probability distributions and thus should ideally be
compared using a metric suitable for distributions, e.g., the
Wasserstein distance. We wish to make use of this metric, but
distance-based classifiers typically 
consider Euclidean distances or inner products in Hilbert spaces only.
To overcome this problem, we designed an embedding that transforms a histogram
into a vector such that the Euclidean distance between two such
vectors approximates the $2$-Wasserstein distance between their
original distributions.

\begin{definition}
 For a probability measure $\mu$ on $\mathbb{R}$, we
 define the \emph{Wasserstein embedding} of dimension $M\in\mathbb{N}$ as
 \[
   \mathcal{E}_M(\mu) = \left(\langle F_{\mu}^{-1},
     \xi_n\rangle\right)_{n=1}^M,
 \]
 where $\{\xi_n\}_{n\in\mathbb{N}}$ is the classical Fourier basis of $L^2(0,1)$.
\end{definition}

$\mathcal{E}_M$ basically computes the first $M$ Fourier coefficients
of the generalized inverse distribution function, also known as the
\emph{quantile function}, $F_{\mu}^{-1}$, cf. \cite{Billingsley}. With
this embedding, we can use the $2$-Wasserstein distance via a Euclidean distance.

\begin{lemma}
 Let $\mu,\nu$ be probability measures on $\mathbb{R}$ with associated quantile
 functions $F_{\mu}^{-1}$ and $F_{\nu}^{-1}$, respectively.
 Assume that $g := F_{\mu}^{-1} - F_{\nu}^{-1}\in L^2(0,1)$, and
 denote by $W_2(\mu,\nu)$ 
 the $2$-Wasserstein distance between the probability measures.
 Then
 \[ \| \mathcal{E}_M(\mu) - \mathcal{E}_M(\nu) \|_2 \to W_2(\mu,\nu) \]
 as $M\to\infty$.
\end{lemma}
\begin{IEEEproof}
In~\cite{StatisticsInWassersteinSpace} it was shown that
\begin{displaymath}
 W_2^2(\mu,\nu) = \int_0^1 \left| F_{\mu}^{-1}(x) - F_{\nu}^{-1}(x)
 \right|^2 \, dx. 
\end{displaymath}
Due to linearity of the inner product we have that
\begin{displaymath}
 \|\mathcal{E}_M(\mu) - \mathcal{E}_M(\nu)\|_2^2
   = \sum_{n=1}^M |\langle g,\xi_n \rangle |^2,
\end{displaymath}
and for $M \to \infty$ we obtain pointwise convergence since
\begin{displaymath}
 \lim_{M\to\infty} \sum_{n=1}^M |\langle g,\xi_n \rangle |^2
  = \sum_{n=1}^{\infty} |\langle g,\xi_n \rangle |^2
  = \|g\|_2^2
  = W_2^2(\mu,\nu)
\end{displaymath}
where we used Parseval's identity.
\end{IEEEproof}
While we specifically chose the classical Fourier basis of~$L^2(0,1)$
by design, we remark that the pointwise convergence result holds for
any orthonormal basis of this space. However, the choice of the
well-known sine/cosine basis has the advantage that estimates can be
made about the rate of convergence provided the function $g$ satisfies
some regularity conditions. For example,
if both generalized inverse distribution functions are bounded, the
Euclidean distance between embeddings converges linearly to the
Wasserstein distance between the original measures. Moreover, as
Fourier approximations are particularly well-suited for smooth
functions even better rates of convergence can be obtained for
embeddings over families of measures with smooth quantile functions. 

\subsection{Support Vector Machines}
With a set of features with geometric relevance at hand, we decided to
employ a Support Vector Machine classifier for the voxelwise 
decision making. SVMs are maximum margin classifiers which aim to
separate two datasets in some feature space in the best possible way by a hyperplane  while simultaneously allowing for some misclassifications during training
in order to increase the robustness of the model,
cf.~\cite{VapnikSVNetworks}.
In our work, we specifically used a variant called
$\nu$-SVM~\cite{LIBSVMPaper}. Denote by $\{(\bm{x}_i,y_i)\}_{i=1}^M$ a
training dataset of samples $\bm{x}_i\in\mathbb{R}^d$ and associated
labels $y_i\in\{\pm 1\}$, $i=1,\ldots,M$. The goal is to find the best
separating hyperplane parametrized by a \emph{normal vector} $\bm{w}$
and an \emph{offset} $b$; finding this hyperplane can be formulated as
an optimization problem 
\begin{align}
  \label{eq:nuDef}
 \min_{\bm{w},b,\bm{\xi},\rho}\; &\frac{1}{2} \|\bm{w}\|_2^2 - \nu
                                   \rho + \frac{1}{M} \sum_{i=1}^M
                                   \xi_i \\
  \nonumber
  s.t.\; &y_i \left(\bm{w}^T\phi(\bm{x}_i)+b\right) \geq \rho-\xi_i, \\
  \nonumber
                                 &\xi_i \geq 0, \; \rho \geq 0,
\end{align}
where $\xi_i$, $i=1,\ldots,M$, are slack variables that allow some
misclassifications during training, and $\phi$ is a function lifting a
sample into feature space. 
We decided to use this variant because it is equivalent to the
commonly used C-SVM~\cite{InterpretationNu} and according
to~\cite{LIBSVMPaper} the hyperparameter $\nu \in (0,1]$  can be
bounded from above by
\begin{equation}\label{eq:numax}
 \nu \leq \nu_{\max} := \frac{2}{M}\,\min \left\lbrace \# \{y_i=+1\}, \#\{y_i=-1\} \right\rbrace \leq 1,
\end{equation}
where $\# \{y_i=\pm1 \}$ denotes the number of samples labeled $+1$
and $-1$, respectively. In the application, we used the well-known
LIBSVM library for solving a SVM optimization problem of the above
form, which actually aims to solve the according dual problem
\begin{align*}
 \min_{\bm{\alpha}}\; &\frac{1}{2}\,\bm{\alpha^T Q \alpha} \\
 s.t.\; &0 \leq \alpha_i \leq \frac{1}{M}, \quad i=1,\ldots,M \\
 &\bm{1^T\alpha} \geq \nu,\;\bm{y^T\alpha = 0},
\end{align*}
where $\bm{Q}_{i,j} = y_i y_j k(\bm{x}_i,\bm{x}_j)$, $i,j=1,\ldots,M$,
is a matrix encoding the kernelization for a kernel $k$. In the
application we use a Gaussian kernel, i.e.,
\begin{equation}
  \label{eq:KernelDef}
  k(\bm{x},\bm{y}) = \langle \phi(\bm{x}), \phi(\bm{y}) \rangle =
  \exp\,\left(-\gamma \|\bm{x}-\bm{y}\|^2\right)    
\end{equation}
with an additional kernel bandwidth parameter $\gamma > 0$.
Support Vector Machines have the special advantage that after the training only few 
samples $\bm{x}_j$, $j=1,\ldots,M'$, where $M' \leq M$, really
contribute to the prediction. These samples are named \emph{support
  vectors} as they lie on the boundary of the margin along the
trained hyperplane. This condition implies that 
$\alpha_j > 0$, $j=1,\ldots,M'$, for these samples.
A trained SVM then can predict a label from its support vector representation by
\begin{displaymath}
 \mathcal{S}^u(\bm{x}) = \sum_{i=1}^M y_i \alpha_i k(\bm{x},\bm{x}_i) + b
\end{displaymath}
and taking the sign of the result.

In its basic form, an SVM only emits a label, which by our problem
formulation is either $+1$ or $-1$. In practice, we prefer a
probabilistic prediction of how likely it is that a given voxel is
part of the component to be segmented and this information shall be
present in the segmentation result.
This type of output is achieved by using Platt
scaling~\cite{PlattScaling} which fits a sigmoidal curve to the
distribution of distances to the separating hyperplane obtained after
training. The goal of Platt scaling is to find a parametrization 
of this sigmoidal curve using logistic regression such that it creates
a calibrated probabilistic estimation $\mathcal{S}(\bm{x}) := P \left(
  y_i = +1 \mid \mathcal{S}^u(\bm{x}) \right)$.
In the sequel, we will refer to such estimates as \emph{confidence values}.

\subsection{Hyperparameter Tuning}
The $\nu$-SVM problem still contains two hyperparameters, i.e.,
parameters occuring in the problem formulation that are not optimized
by the solver itself but have to be tuned manually.
In our formulation, these remaining variables are the value
  $\nu$ in~\eqref{eq:nuDef} that determines how tolerant the SVM is towards
  misclassifications and the localization parameter $\gamma$ in the
  kernel function $k$ in~\eqref{eq:KernelDef}.
In~\cite{InterpretationNu}, it was shown that $\nu$ is ``\textsl{an upper bound
  on the fraction of margin errors and a lower bound of the fraction of
  support vectors}'' and that these two fractions coincide with probability
$1$. Therefore, $\nu$ effectively serves as a trade-off between
misclassifications and sparsity of the model. 
Clearly, the value of $\nu$ is contained in the unit interval, and we
already mentioned that it can be bounded from above by $\nu_{\max}$ as
given in Equation~\eqref{eq:numax}.
In order to find a good value for this hyperparameter, we sample the
feasible region $(0,\nu_{\max}]$ equidistantly.
On the other hand, $\gamma > 0$ is the kernel bandwidth parameter
inversely proportional to the variance of the Gaussian kernel
function. As is common practice, we create for it a parameter set on a
logarithmic scale. The boundaries for both sets in the real
applications were determined experimentally. 
Then, the overall hyperparameter grid is given as $P = P_{\nu} \times P_{\gamma}$.
From this, we determine the best hyperparameter grid
point~$(\nu^*,\gamma^*)\in P$ using $C$-fold cross validation where
$C=7$ turned out to be suitable in our applications.

\subsection{Why Support Vector Machines?}
In principle, the classifier need not be a Support Vector Machine and
one could try more fancy techniques of machine learning, so we briefly
want to state why we decided to choose an SVM.
First of all, its most important advantage is that only a small number
of data points is necessary to make a prediction, which saves both
runtime as well as memory and makes SVMs especially well-suited for
our active learning approach: a robust hyperplane can be trained with
few samples, i.e., with a small number of seed voxels. This keeps the
amount of user interaction very moderate.
Another advantage is that we could encode our geometric knowledge
about local regions in our features,
which are thus designed such that visually similar regions produce
feature vectors that are close with respect to the
Euclidean distance and thus pair well with our choice of a Gaussian
kernel. The nonlinearity in the Gaussian kernel enables the  
construction of complex separation surfaces which still fit into the
hyperplane framework in the kernel space.

Furthermore, support Vector Machines are known to provide a good
out-of-sample generalization, i.e., bias contained in the training
dataset can be somewhat compensated by appropriate hyperparamter
tuning~\cite{SVMAdvantages3}.
Finally, the low memory consumption makes SVMs attractive to be
serialized and applied to other datasets. 
Although in practice a pretrained model will rarely be a good
classifier for an unrelated and unseen dataset, a simple retraining
method can be formulated by serializing all unscaled feature
vectors. Upon retraining, the serialized representation is loaded from
disk and combined with the newly provided samples which forms the new
training data. This simple and straightforward approach gave more than
satisfactory results in many of our applications.

\subsection{Active Learning and Segmentation}
After designing geometric features based on local voxel environments
and a classifier which can be trained on them, we put everything
together into a segmentation procedure based on \emph{active
  learning}.

Active learning (AL) is an emerging branch in machine learning that
tries to compensate a lack of annotated training data by featuring an
oracle which labels samples on-the-fly. Examples for such oracles are
additional machine learning techniques or in our and many other
applications a human annotator.
The basic concept is that some samples are labeled by the oracle, and
a model is trained from them. This trained model then predicts results
which render some samples as ambiguous, i.e., the model cannot make a
clear prediction about this sample. The ambiguous samples are then
returned to the oracle to be labeled.
This process is continued until the results are stable or good enough.
There exist different approaches for querying the oracle with samples
that are either generated by the model \emph{de novo} or stem from a
stream or a pool of existing samples. 
In the literature, a variety of techniques for detecting ambiguous
samples can be found, including
\emph{uncertainty sampling}, in which an uncertainty measure
determines the samples which the model is most uncertain about,
\emph{query-by-committee}, where the samples are selected on which the
majority of the classifiers disagree, or
\emph{estimated error reduction} which selects the sample that minimizes the 
generalization error of the model, cf.~\cite{ActiveLearningSurvey}.

Applied to voxelwise segmentation, active learning would imply that
our model queries a user to label small local voxel environments. As
this is impractical, we instead follow a pool-based approach in which
a trained model segments the entire voxel volume.
  Using a priori
  information this could be restricted to selected regions only, however
  the definition and handling of such information is highly
  application specific and out of scope of our work here.
Next, ambiguous samples are detected by uncertainty sampling, computed from
our confidence value predictions. In our setting of segmenting very
large three-dimensional volumes, more advanced sample query
strategies, e.g., estimated error reduction, are too computationally
expensive. 
To perform uncertainty sampling, we compute an uncertainty value from
a confidence value as follows.
\begin{definition}\label{def:uncertainty}
  For a \emph{confidence value} $\mathcal{S}(\bm{x}) \in [0,1]$ and
  $\delta \geq 0$, we define the \emph{uncertainty value}
  $\mathcal{U}(\bm{x})$ as
  \[
    \mathcal{U}(\bm{x}) 
    = \exp\,\left( 
      -\delta \tan\left( 
        \pi \left| \mathcal{S}(\bm{x}) - \frac{1}{2} \right|
      \right)
    \right).
  \]
\end{definition}
By this definition of the confidence values, we know that the model is
most uncertain about samples which lie close to the separating
hyperplane and thus produce a confidence value close to
$1/2$. The uncertainty values define a Gaussian window around
this confidence value where $\delta$ controls the width of that bell
curve.

The segmentation procedure is repeated until the results are of
sufficient quality, and it is summarized in 
Algorithm~\ref{alg:segalg}.
\begin{figure}[!t]
 \removelatexerror
 \begin{algorithm}[H]
  \caption{AL Volume Segmentation}
  \label{alg:segalg}
  \DontPrintSemicolon
  \SetKwInOut{Input}{Input}
  \SetKwInOut{Output}{Output}
  \Input{Input volume $V$}
  \Input{Feature computation function $F$}
  \Output{Output volume $\widetilde{V}$}
  \BlankLine
  $(f_i, L_i) \gets $ de-serializ.~\tcp*{opt.,~default~$((), ())$}
  $U \gets V$\;
  \Repeat{$\widetilde{V}$ is satisfactory}{
   $(f_n, L_n) \gets $ features and labels, selected using $U$\;
   $f_i \gets (f_i, f_n)$\;
   $L_i \gets (L_i, L_n)$\;
   Train SVM using features $f_i$ and labels $L_i$\;
   Serialize trained model \tcp*{opt.}
   \ForAll{$\bm{\alpha}\leq\bm{d}$}{
     $E \gets $ $K\times K\times K$ environment around $\bm{\alpha}$\;
     $\widetilde{x}_{\bm{\alpha}} \gets \lfloor 100\,\mathcal{S}(F(E)) \rfloor$\;
   }
   $U \gets $ uncertainty volume from confidence values\;
  }
 \end{algorithm}
\end{figure}
The framework is a follows: We start with an input volume $V$ of
dimensions $\bm{d}\in\mathbb{N}^3$ that is assumed to consist of
voxels located at the regular \emph{grid positions}
$\bm{\alpha}\leq\bm{d}$ with values $x_{\bm{\alpha}}\geq 0$. 
The generated output volume $\widetilde{V}$ is of same size as $V$ and
its voxel values are denoted by $\widetilde{x}_{\bm{\alpha}}$.
In a first (optional) step, serialized feature vectors and associated
labels can be read from disk.
Next, an oracle selects voxels of high uncertainty and labels them
either positively ($+1$ = ``belongs
to the relevant component'') or negatively ($-1$ = ``does not belong
not to the component''), where the selection is focussed on visual
input from the original voxel volume. 
In subsequent iterations, the uncertainty volume $U$
hints about regions where the model is unsure and from which the
oracle can select further samples. Around each selected sample a small
local voxel environment is extracted from which features $f_n$ are computed.
Then a Support Vector Machine is trained using all feature vectors and
their labels. This step also includes training the feature vector
scaler, the hyperparameter tuning as well as Platt scaling. 
After this training is completed, all voxels for which a full local
region can be extracted are investigated in a single pass, where,
again, a small voxel environment $E$ is extracted around each of them, and
features are computed from which the SVM predicts 
a confidence value $\mathcal{S}(F(E))$ which is scaled to percent.
From this confidence, an uncertainty volume can be computed according
to Definition~\ref{def:uncertainty}, which hints to the oracle where
to choose the next samples. This process is repeated until the results
are satisfactory.

\begin{figure*}[!ht]
 \centering
 \subfloat[Iteration \#1]{\includegraphics[clip,trim=0 4mm 0 5mm, width=.48\textwidth]{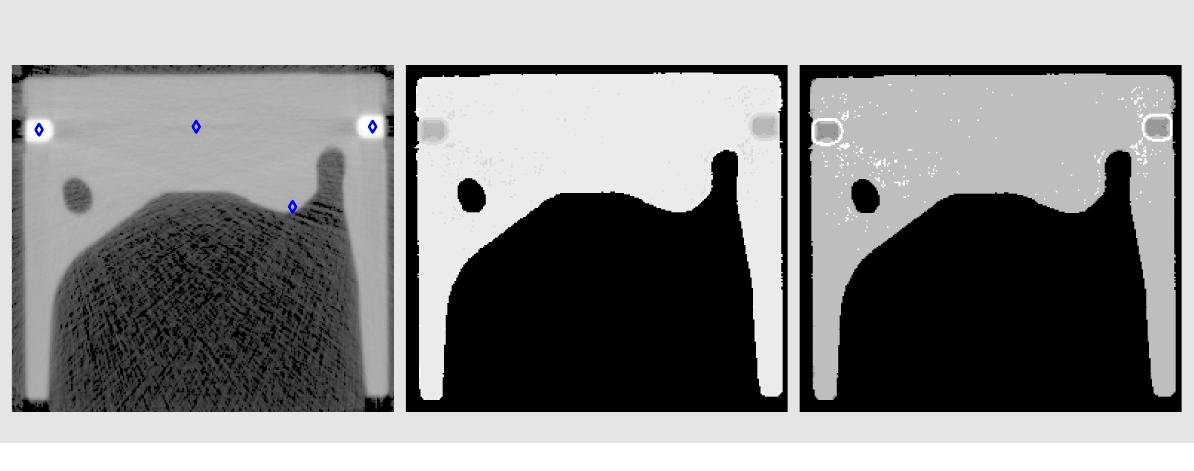}\label{fig:a1}} \hfil
 \subfloat[Iteration \#2]{\includegraphics[clip,trim=0 4mm 0 5mm, width=.48\textwidth]{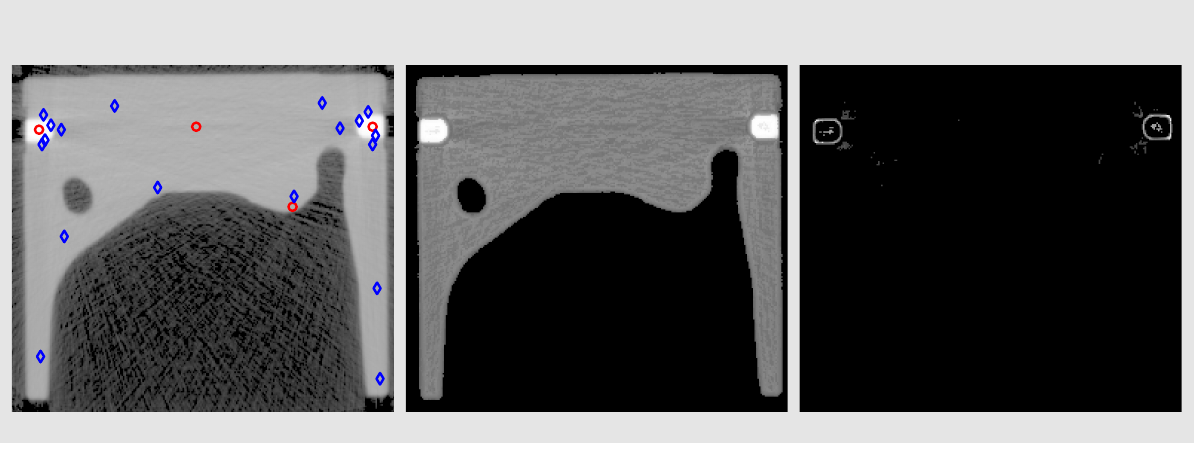}\label{fig:a2}} \hfil
 \subfloat[Iteration \#3]{\includegraphics[clip,trim=0 4mm 0 5mm, width=.48\textwidth]{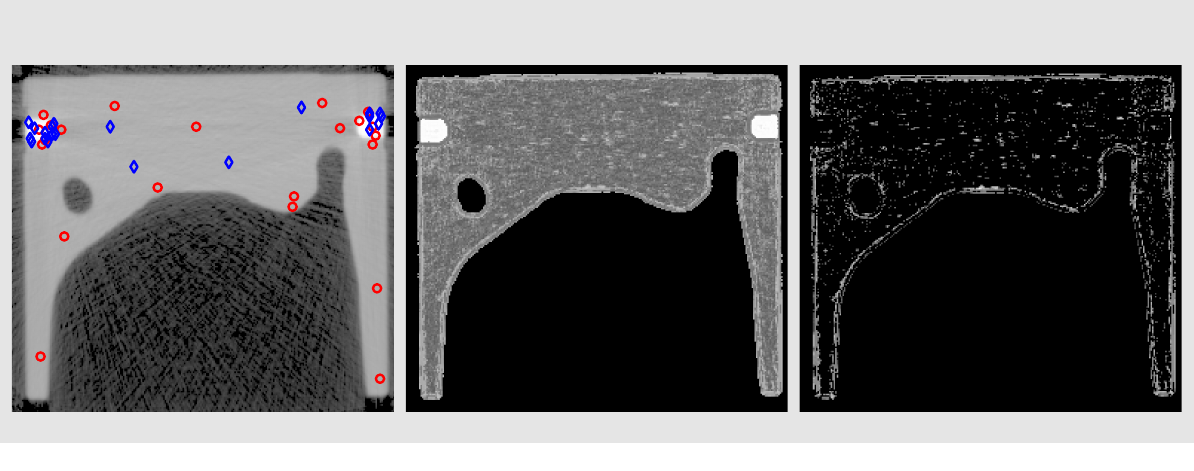}\label{fig:a3}} \hfil
 \subfloat[Iteration \#4]{\includegraphics[clip,trim=0 4mm 0 5mm, width=.48\textwidth]{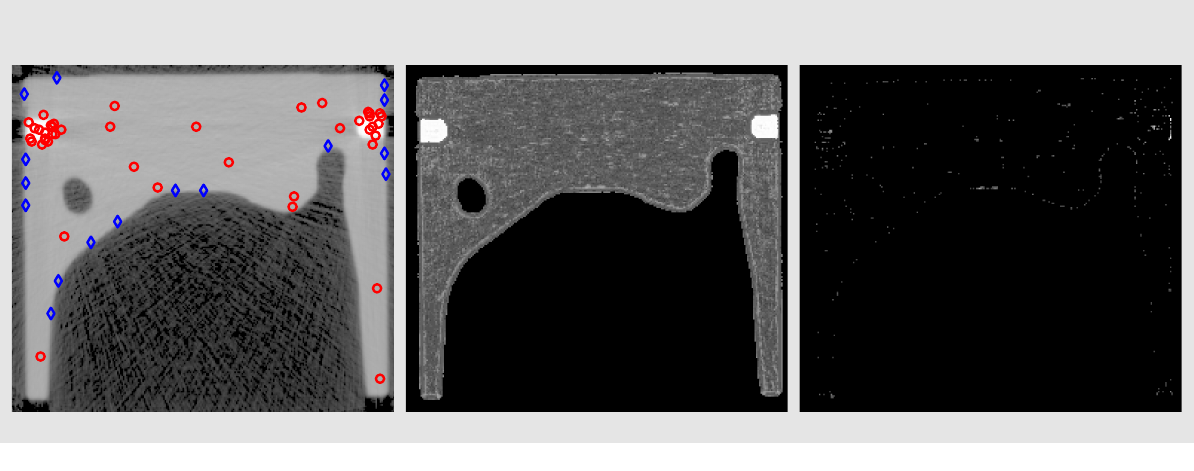}\label{fig:a4}} \hfil
 \caption{Applying Algorithm~\ref{alg:segalg} to a motor piston dataset, showing four iterations. 
          In all images contrast was enhanced for better visibility, brighter shades indicate
          higher values. \protect\subref{fig:a1} shows the setting of the first iteration in which first
          the voxel selection is shown, new ones by blue diamonds and already selected ones by red circles.
          Its middle image depicts the predicted confidence values and to its right the computed uncertainty
          values are visualized. This is best viewed in color.
          \protect\subref{fig:a2},~\protect\subref{fig:a3} and~\protect\subref{fig:a4} show the next three
          iterations, respectively.%
 }
 \label{fig:alsegexample}
\end{figure*}

\medskip
To illustrate the interactive procedure we consider a CT scan of a
Mahle motor piston and aim to segment the steel ring located near the
top. The process took four iterations, the intermediate results are
depicted in Figure~\ref{fig:alsegexample} in which brighter colors
indicate higher values of the CT grayscale.
In the initial iteration, we interactively selected \emph{two} seed
voxels in the steel ring regions to be marked as
positive and additionally chose two seed voxels in the aluminum
regions of the piston, marked as negative.
After applying the segmentation, we then obtain a volume consisting of
confidence values from which we compute uncertainty volumes. Both the
confidence and uncertainty values in the images shown in Figure~\ref{fig:a1}
indicate that the model is highly unsure about decision making.

We next use these uncertainty values as an indication where to select
additional seed voxels and apply  Algorithm~\ref{alg:segalg} again,
see Figure~\ref{fig:a2}.
The results improve by a magnitude as now confidences are high at the
steel ring while they are low for the other piston voxels. The
uncertainty is only concentrated at the border of the datasets to
separate. We continue this for two further iterations during which the
confidence values stabilize and the uncertainty decreases.

After four iterations, the uncertainty is mostly gone, thus signaling
that we can now stop selecting voxels. A simple thresholding on the
confidence values then yields the desired segmentation result. 

\subsection{Runtime and Memory Analysis}
Since we especially focus on processing large volumetric data, we
briefly analyze the asymptotic runtime and memory behavior.
Reading serialized data from disk is optional and can be considered as
an operation with a constant runtime. 
The next step selects $M\in\mathbb{N}$ seed voxels and computes
feature vectors from local environments of the seed voxels, which are
used for training a Support Vector Machine. 
The training procedure for a SVM is reported to be typically of
quadratic or cubic order, depending on its hyperparameters
cf.~\cite[p.10]{SVMCubic}. Our training step includes a grid search 
hyperparameter tuning where for each parameter a $C$-fold cross
validation is executed. 
Therefore, the overall runtime necessary for our training step scales
with $\#P \, C \, \poly(M/C)$.
After the training is completed, we iterate over the volume in a
single linear pass and perform the voxelwise classification. That step
is linear in terms if the number $N\in\mathbb{N}$ of voxels.
Note that the computation of the uncertainty volume may be not be
executed at all if, e.g., an 
application decides to simply use the confidence values directly for
indication where to select voxels next. If it is used, it is again
linear in $N$.
That interactive procedure is repeated until the result is
satisfactory after $I\in\mathbb{N}$ iterations. Again, the optional
final serialization step has a constant runtime.
In summary, the overall runtime behavior of Algorithm~\ref{alg:segalg} is
\[
  \mathcal{O}\left( I \left( \frac{\#P}{N_T}\, C\, \poly(M/C) + N \right) \right),
\]
where typically $\deg\,\poly(M/C) \leq 3$ and $N_T$ denotes the number
of threads available on the executing system since the grid search
trainings can be executed independently from another.
Since $C$, $P$ and $N_T$ are fixed a priori and since we expect the
number $M$ of seed voxels  
to be typically far smaller (typically by orders of magnitude) and
\emph{independent} of $N$, the runtime is effectively  
linear in the number of iterations and the number of voxels, even in
our pessimistic worst case estimate.
This makes our procedure especially well-suited for processing very
large volumetric datasets where only asymptotically linear algorithms
w.r.t. to $N$ are considered applicable. 

Regarding memory consumption, we note that our segmentation procedure
first needs to compute the feature vectors of dimensionality $d$ for
all $M$ seed voxels and needs to keep these in memory at all times.
The memory usage during training the model is roughly dominated by
computing the kernel matrix of the dual problem formulation, which is
naively of size $M^2$.
While implementation issues might improve upon this, e.g., using a
sparse matrix representation if possible, the dominating factor is
still quadratic in the number of seed voxels. 
During the voxelwise classification only local environments centered
around the currently processed voxel and the trained model are
necessary, i.e., we need the $M$ trained 
weight coefficients and the local environment of constant size.
In total, the memory consumption is of order 
$\mathcal{O}\left( M^2 + M d + M \right)$.
Since $M$ is again expected to be small and
independent of the number of voxels, our approach is well-suited even
for segmentation of large voxel volumes which no longer fit into the main
memory of a computer.

\subsection{Multiresolution Segmentation}
By design, our method is effectively a linear pass over the input
volume which makes it applicable to volumes without size restrictions
while at the same time being as efficient as possible for 
voxelwise segmentation. However, its execution time clearly increases
linearly in the volume size which means cubically in the resolution.
However, in practice the components to be found in and segmented from
the volume are often relatively small compared to the overall
dataset. Thus, the idea of multiresolution processing arises
naturally.

While some related work considers a multiscale hierarchy to extract
features and perform a classification on the full dataset,
cf.~\cite{MR1,MR3}, our focus lies on segmenting only \emph{relevant}
regions. 
In a similar spirit, \cite{MR2} segments image stacks based on a
Wavelet transform using an expectation-maximization-based 
approach. Specifically for 3D volumes, \cite{MR4} proposed a
segmentation method on the coarsest level of a Gaussian pyramid and then
refines the results on finer levels. In connection with interactive
segmentation very few papers were published of which we mention~\cite{MR5} and~\cite{MR6}.
\cite{MR5} proposes an interactive shape
model which uses features extracted 
across multiple resolutions but processes the data in its original
format. Closest to our approach is the work in \cite{MR6},
where retinal image data is segmented progressively using simple color
information. 

Unfortunately, common hierarchical representations need be serialized to disk,
which is exactly what we want to avoid here. Instead, we implemented a
local multiresolution framework in which each voxel value currently
necessary is computed ad hoc. 
To describe our apporoach in detail, let $V^{\ell'}$ be a volume of dimensions
$\bm{d}\in\mathbb{N}^3$ at resolution level
$\ell'\in\{1,\ldots,\ell_{\max}\}$, where $\ell_{\max} \in\mathbb{N}$
denotes the number of resolution levels available for this volume. 
Assume that it consist of voxels located at a grid
$\bm{\beta}\leq\bm{d}$ having values $x_{\bm{\beta}}$.
We now wish to interpret such a volume as a volume at a coarser
resolution level $\ell\in\{1,\ldots,\ell'\}$. To that end, a voxel at
position $\bm{\alpha}$ in the coarse grid is said to \emph{cover} a
set of voxels in the fine grid which are combined for its
computation. Formally, we define the set of its covered voxel
positions by 
\begin{displaymath}
  \mathfrak{S}_{\ell'}^{\ell}(\bm{\alpha})
  = \left\lbrace
    \bm{\beta} \leq \bm{d} \bigm\vert 
    \bm{\beta} = 2^{\ell'-\ell} \bm{\alpha} + \bm{\kappa}, \; |\bm{\kappa}| \leq 2^{\ell'-\ell} - 1
  \right\rbrace
\end{displaymath}
which encompasses $8^{\ell'-\ell}$ high-resolution voxels in total.
Like in Haar wavelet computations, the grayscale value of any voxel at
position $\bm{\alpha} \leq 2^{\ell'-\ell} \bm{d}$ on a coarser
resolution level  
is then computed as the arithmetic mean of the voxel values covered by it:
\begin{displaymath}
  x_{\bm{\alpha}} = 8^{\ell-\ell'}
  \sum_{\bm{\beta}\in\mathfrak{S}_{\ell'}^{\ell}(\bm{\alpha})}
  x_{\bm{\beta}}. 
\end{displaymath}
Typically, we perform this on-the-fly computation at coarse
resolutions $\ell$ directly from the finest resolution level, i.e.,
the actual grayscale data, thus in our application we work with $\ell' = \ell_{\max}$.

Based on these \emph{multiresolution volumes}, we can now formulate an
enhanced segmentation procedure that combines the multiresolution
interpretation as well as our active learning procedure. 
The result is listed in Algorithm~\ref{alg:mrseg}.
\begin{figure}[!t]
 \removelatexerror
 \begin{algorithm}[H]
  \caption{Multiresolution segmentation}
  \label{alg:mrseg}
  \DontPrintSemicolon
  \SetKwInOut{Input}{Input}
  \SetKwInOut{Output}{Output}
  \SetKwFunction{FindCandidates}{FindCandidates}
  \Input{Input volume $V$ of dimensions $\bm{d}$}
  \Input{Feature computation function $F$}
  \Output{Output volume $\widetilde{V}$}
  \BlankLine
  $(f_i, L_i) \gets $ de-serializ.~\tcp*{opt.,~default~$((), ())$}
  $U \gets V$\;
  \Repeat{$\widetilde{V}$ is satisfactory}{
   $(f_n, L_n) \gets $ features and labels, selected using $U$\;
   $f_i \gets (f_i, f_n)$\;
   $L_i \gets (L_i, L_n)$\;
   Train SVMs $m_{\ell}, \ell=1,\ldots,\ell_{\max}$\;
   Serialize trained models \tcp*{opt.}
   $(\rho_{\ell})_{\ell=1}^{\ell_{\max}} \gets $ confidence thresholds\;
   \BlankLine
   \tcp{Region localization}
   $C_0 \gets \{(\bm{0},\bm{d})\}$\;
   \For{$\ell \gets 1 \to \ell_{\max}-1$}{%
     $C_{\ell} \gets $ \FindCandidates{$V^{\ell},C_{\ell-1},m_{\ell},\rho_{\ell}$}\;
   }
   \BlankLine
   \tcp{Final voxelwise segmentation}
   \ForAll{$C \in C_{\ell_{\max}-1}$}{
     $V_C \gets $ voxel positions in $C$\;
     \ForAll{$\bm{\alpha}\in V_C$}{
       $E \gets $ $K\times K\times K$ environment around $\bm{\alpha}$\;
       $\widetilde{x}_{\bm{\alpha}} \gets $ SVM prediction $m_{\ell_{\max}}(F(E))$\;
     }
   }
   \BlankLine
   $U \gets $ uncertainty volume from confidence values\;
  }
 \end{algorithm}
\end{figure}
Also this procedure requests the labeling of additional seed
voxels from which features are extracted. 
That feature extraction now happens on all resolution levels
\emph{simultaneously} and one specific SVM is trained for each
scale. Additionally, confidence thresholds are computed which
determine the minimal value to consider a voxel as a candidate. That
is, during the iterations over all but the finest resolution level,
the SVM related to the current resolution predicts a confidence value for any
voxel in the considered regions at that resolution.
If the confidence is higher than the threshold,
that voxel does likely belong to the component we aim for.
In our application, we search for an optimal threshold which minimizes
the misclassifications, i.e., 
\begin{align*}
\rho_{\ell} =
 \argmin_{\rho\in[0,1]}
  \sum_{\epsilon\in\{\pm 1\}}
   \frac{1}{\#\mathcal{T}_{\epsilon}^{\ell}}
    \sum_{\bm{x}\in\mathcal{T}_{\epsilon}^{\ell}} H\left(\epsilon\left(\rho-\mathcal{S}^{\ell}(\bm{x})\right)\right)
\end{align*}
where $\mathcal{T}_{\epsilon}^{\ell}$, $\epsilon\in\{\pm 1\}$, denotes
the positive and negative training samples at resolution $\ell$,
respectively, and $H$ is the Heaviside step function.
A procedure \texttt{FindCandidates} uses this approach by considering
all voxels of $V^{\ell}$ contained in the previous 
set of candidates $C_{\ell-1}$ on resolution level $\ell-1$. For each
of them, the trained model $m_{\ell}$ predicts a confidence value and 
the currently processed voxel is remembered as a candidate if and only
if that prediction lies above the  
confidence threshold $\rho_{\ell}$. The set of candidate voxels is
simplified periodically by organizing them into
rectangular regions of interest by connectedness and scaling them to
fit the volume dimensions at the next resolution level. After
processing a resolution level, the candidates obtained on that level
serve as the new search locations on the next scale, and we continue
until the finest resolution level on which  
our usual voxelwise classification takes place inside the final
candidate regions.
  The goal is, of course, that the candidate regions encompass a relatively
  small subset of the regions at higher resolutions, thus reducing the
  number of voxels to be considered significantly.
Having processed the volume this way over all resolution levels, we
again may compute an uncertainty volume and repeat the method until
the result is satisfactory. 

The purely local processing that is present also in the multiscale
case, renders Algorithm~\ref{alg:mrseg} applicable to volumes 
without size restrictions too. Roughly, the runtime complexity of the
modified algorithm depends linearly on the 
number of seed voxels, the runtime for the training procedure and the
number of voxels inside the candidate 
regions found. The multiresolution procedure is intended to be used
when we aim to segment relatively 
small parts from really big voxel scans. Naturally, if the object to
be segmented fills the entire scan, a multiresolution 
voxelwise segmentation is not beneficial and the runtime deteriorates
to be polynomial in the number of voxels. 
The memory consumption simply duplicates the ones
derived for the single SVM case by $\ell_{\max}$ and 
thus still scales linearly with the number of seed voxels.

In an experiment, we segmented the springs from a computed tomography
scan of a crashed car which consists of 3,171,818,496 voxels in
total. The remaining regions of interest at the finest resolution
level encompassed 6,281,221 voxels which amounts to roughly 0.2
percent of the overall voxel count, which demonstrates the dramatic
reduction of voxels to process if the dataset is well suited for
multiresolution processing.

\section{Results}\label{sec:results}
We now illustrate the potential and the performance of our approach by
means of some concrete examples.
We provide a quantitative evaluation on scans where ground truth data
is available, but also demonstrate the flexibility of our method on complex
scans from a variety of different domains.
All datasets considered were created by the Fraunhofer Development
Center for X-ray Technology (EZRT) in F\"urth. If necessary, additional
copyright information is provided. 

\subsection{Quantitative Evaluation}
To provide quantitative results, we apply our method in a deterministic
way to seven selected datasets 
where ground truth data was available.
These datasets are depicted  
in Figure~\ref{fig:quantitativedatasets} and
Figure~\ref{fig:fiesta}, respectively.
{
For each of them, we selected a fixed set of seed voxels and the parametrization
shown in Table~\ref{tbl:experimentsetup}, including the features we
used and the size $K_1$ of the local environments. 
For each dataset, we also employed a postprocessing
step in form of a speckle removal, where the values of any
voxels having less than $\eta$ nonzero neighbor voxels in an
environment of size $K_2\times K_2\times K_2$ are set to zero, 
followed by a simple connected components analysis to select the
components we are interested in.
  These straightforward image processing operations were
  needed to clean up the segmentation result by removing isolated
  voxels that were an obvious misclassification.
The results are summarized in Table~\ref{tbl:experimentresults} where
we compared our segmentation results with the ground truth according
to the well-known performance metrics Intersection over Union (IoU),
precision, recall, and the $F_1$ score. 
These experiments clearly show that our very general segmentation method is
able to achieve good results on scans which stem from the industrial
domain.

\begin{table}[!h]
 \renewcommand{\arraystretch}{1.3}
 \caption{Experiment setup for quantitative evaluation.}
 \label{tbl:experimentsetup}
 \centering
 \begin{tabular}{|l||c|l|c|}
  \hline 
  \textbf{Scan} & $K_1$ & \textbf{Features used} & \textbf{Speckle removal} \\ \hline \hline 
  Ford Fiesta& 7 & Grayscale + Inertia     & $K_2=5, \eta=15$ \\ \hline
  Piston     & 5 & Grayscale + Inertia     & $K_2=5, \eta=18$ \\ \hline
  Big Piston & 5 & Grayscale               & $K_2=5, \eta=18$ \\ \hline
  Tasterwald & 7 & Grayscale + LBP         & -                \\ \hline
  Ring       & 7 & Grayscale + Dist. Hist. & -                \\ \hline
  Rumpf      & 7 & Grayscale + Plane fit   & -                \\ \hline
  TPA        & 3 & Grayscale + Inertia     & $K_2=3, \eta=18$ \\ \hline
 \end{tabular}
\end{table}

\begin{table}[!h]
 \renewcommand{\arraystretch}{1.3}
 \caption{Quantitative evaluation results.}
 \label{tbl:experimentresults}
 \centering
 \begin{tabular}{|l||c|c|c|c|}
  \hline 
  \textbf{Scan} & \textbf{IoU} & \textbf{Precision} & \textbf{Recall} & $F_1$ \\ \hline
  Ford Fiesta& $0.757$ & $0.970$ & $0.775$ & $0.862$ \\ \hline
  Piston     & $0.980$ & $0.980$ & $1.000$ & $0.990$ \\ \hline
  Big Piston & $0.985$ & $0.980$ & $0.997$ & $0.992$ \\ \hline
  Tasterwald & $0.883$ & $0.999$ & $0.884$ & $0.934$ \\ \hline
  Ring       & $0.950$ & $0.965$ & $0.984$ & $0.974$ \\ \hline
  Rumpf      & $0.949$ & $0.963$ & $0.985$ & $0.974$ \\ \hline
  TPA        & $0.989$ & $0.992$ & $0.997$ & $0.994$ \\ \hline
 \end{tabular}
\end{table}

\begin{figure*}[!t]
 \centering
 \subfloat[Piston]{\includegraphics[width=.132\textwidth]{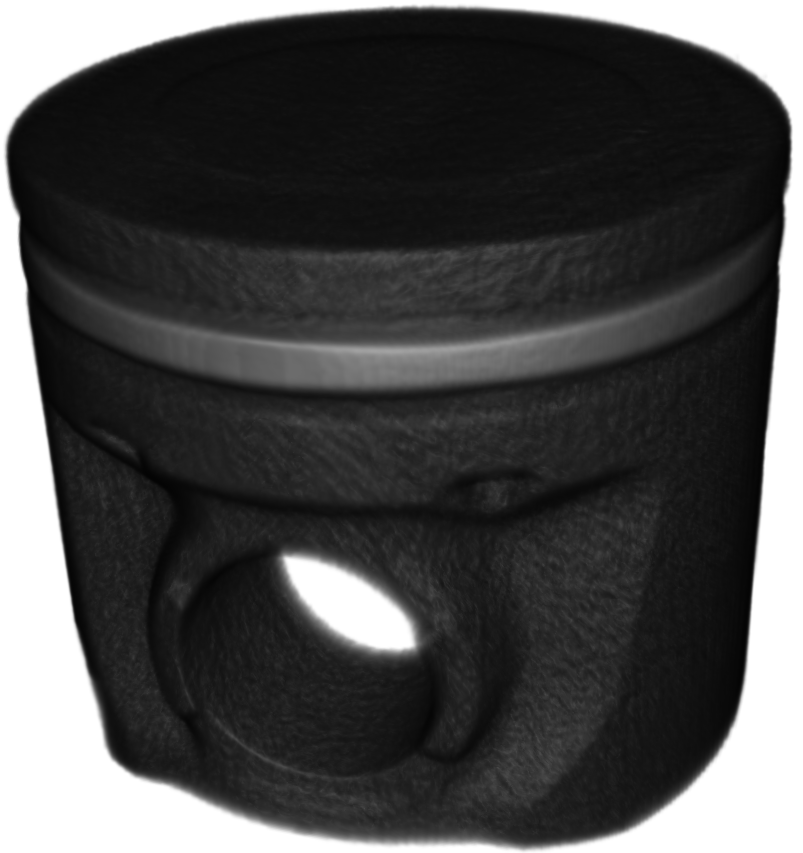}\label{fig:piston}} \hfil
 \subfloat[Big Piston]{\includegraphics[width=.180\textwidth]{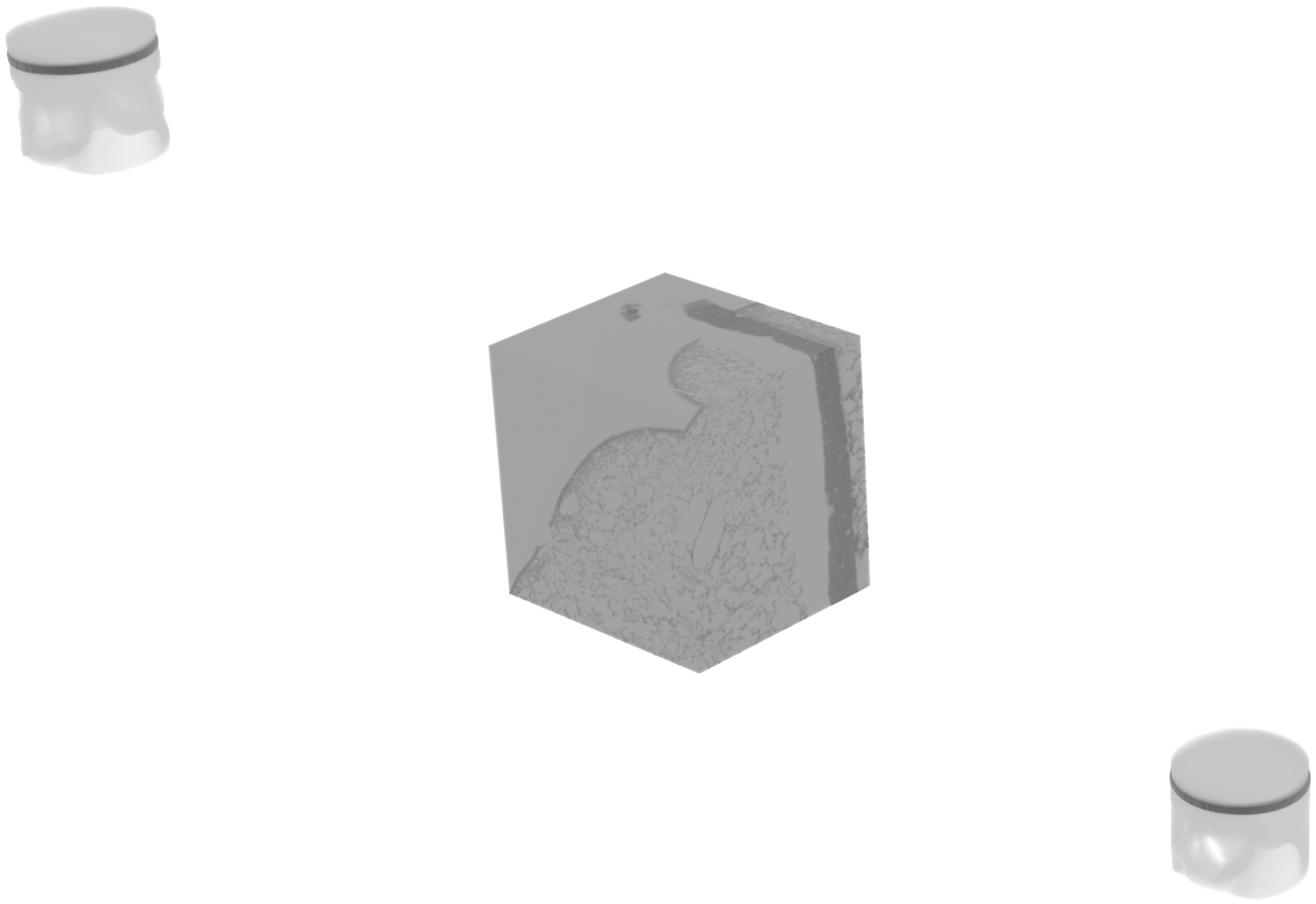}\label{fig:bigpiston}} \hfil
 \subfloat[Tasterwald]{\includegraphics[width=.162\textwidth]{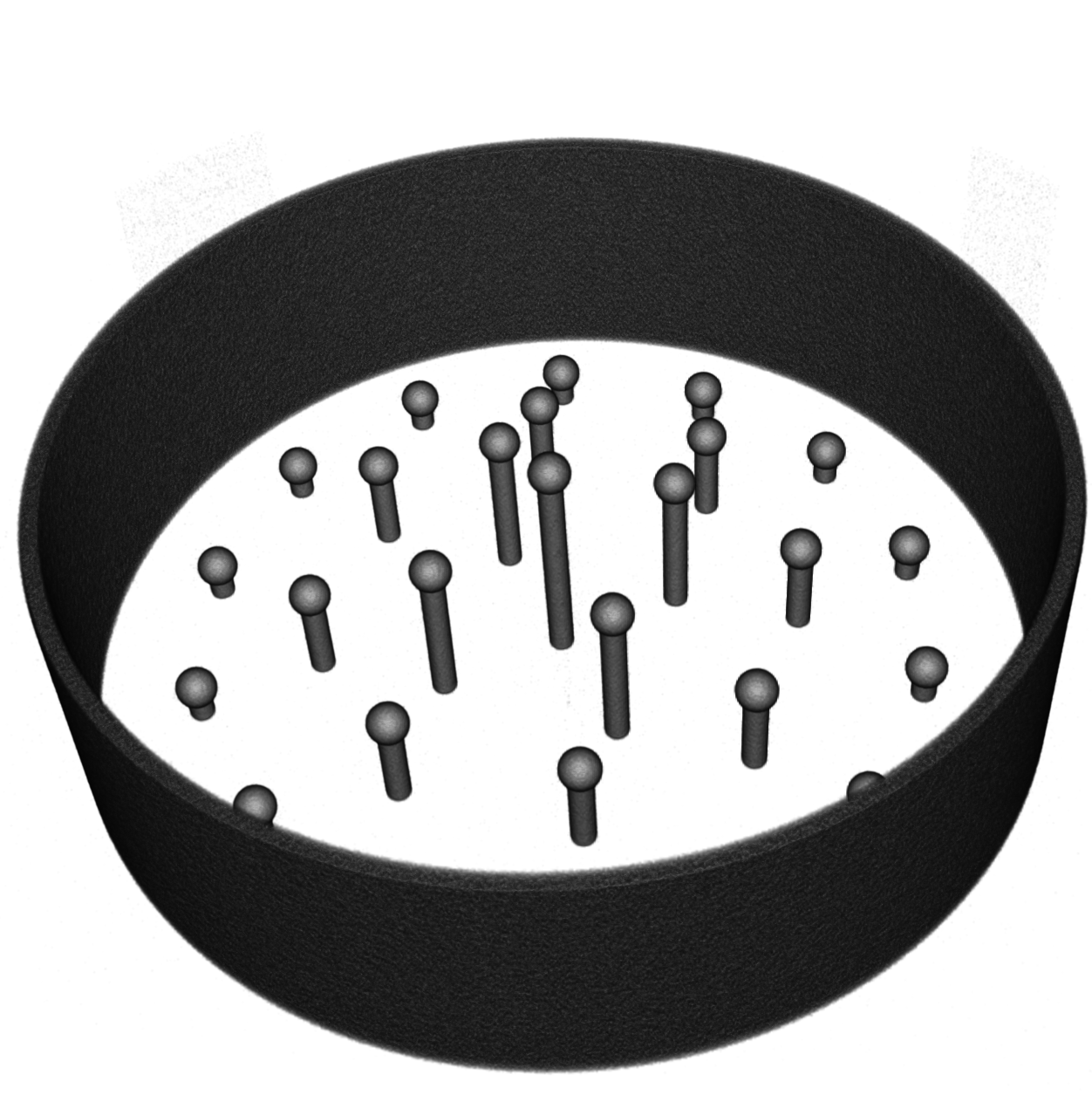}\label{fig:tasterw}}     \hfil
 \subfloat[Ring]{\includegraphics[width=.162\textwidth]{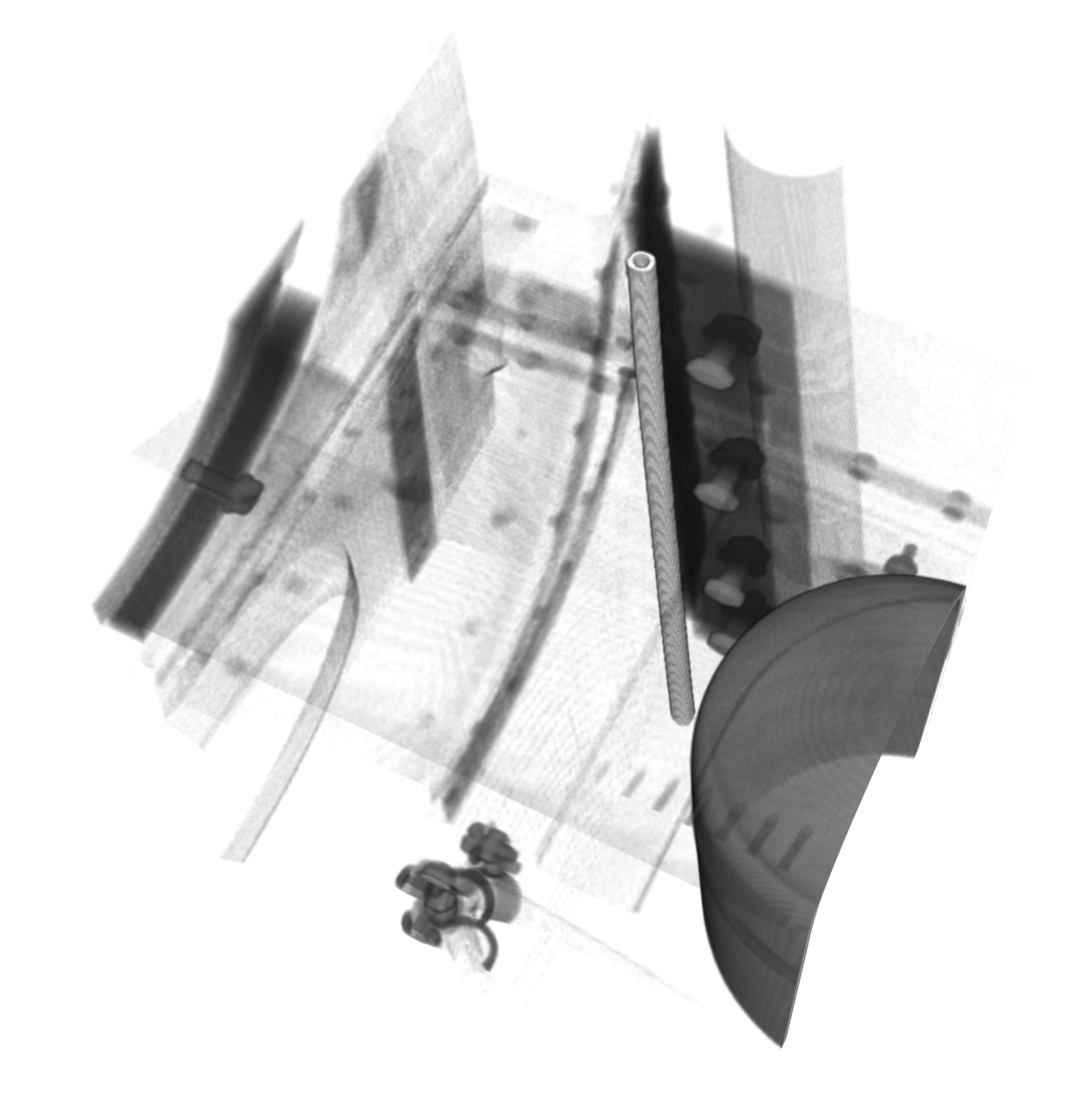}\label{fig:ring}}\hfil
 \subfloat[Rumpf]{\includegraphics[width=.162\textwidth]{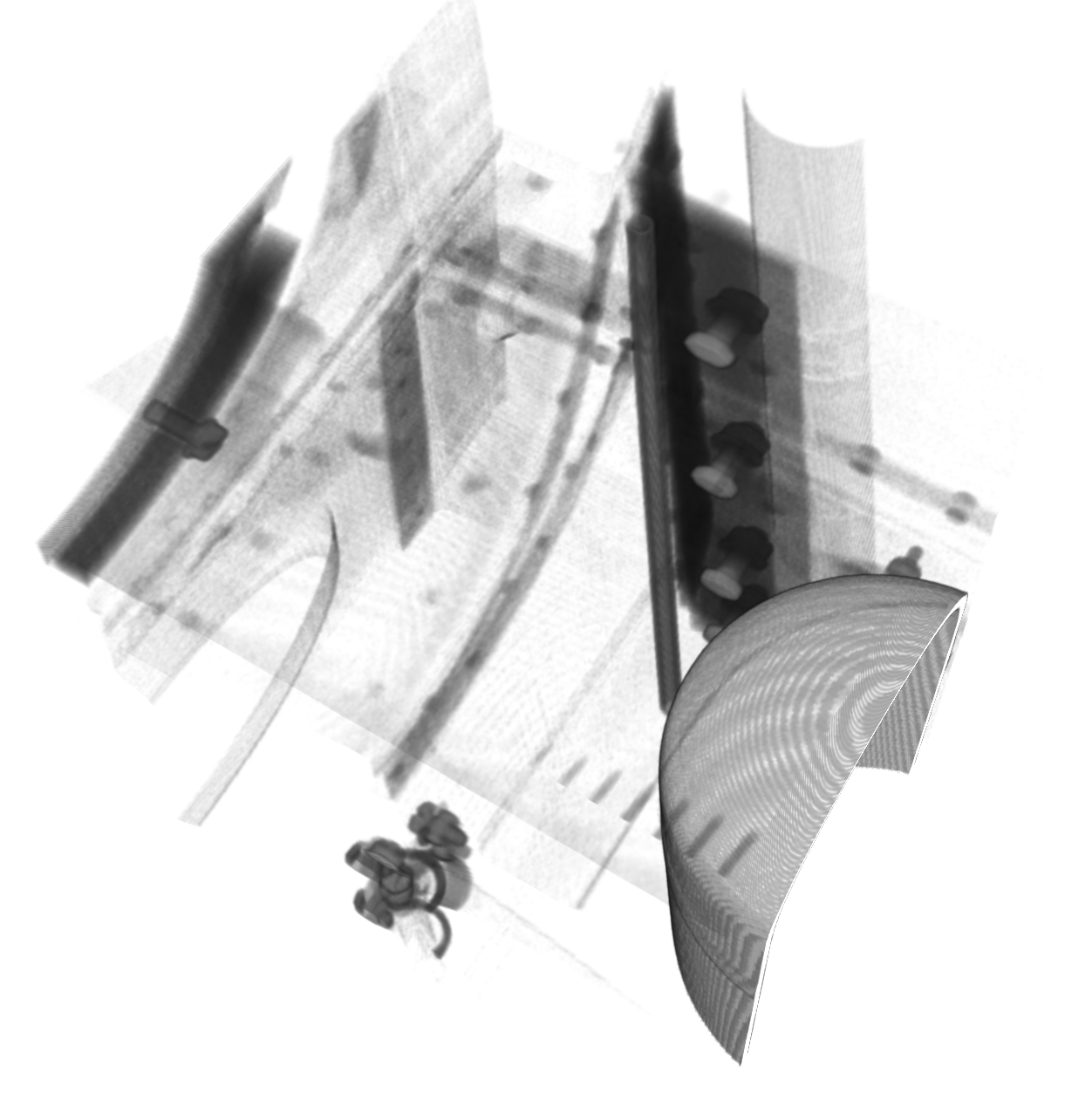}\label{fig:rum}}\hfil
 \subfloat[TPA]{\includegraphics[width=.162\textwidth]{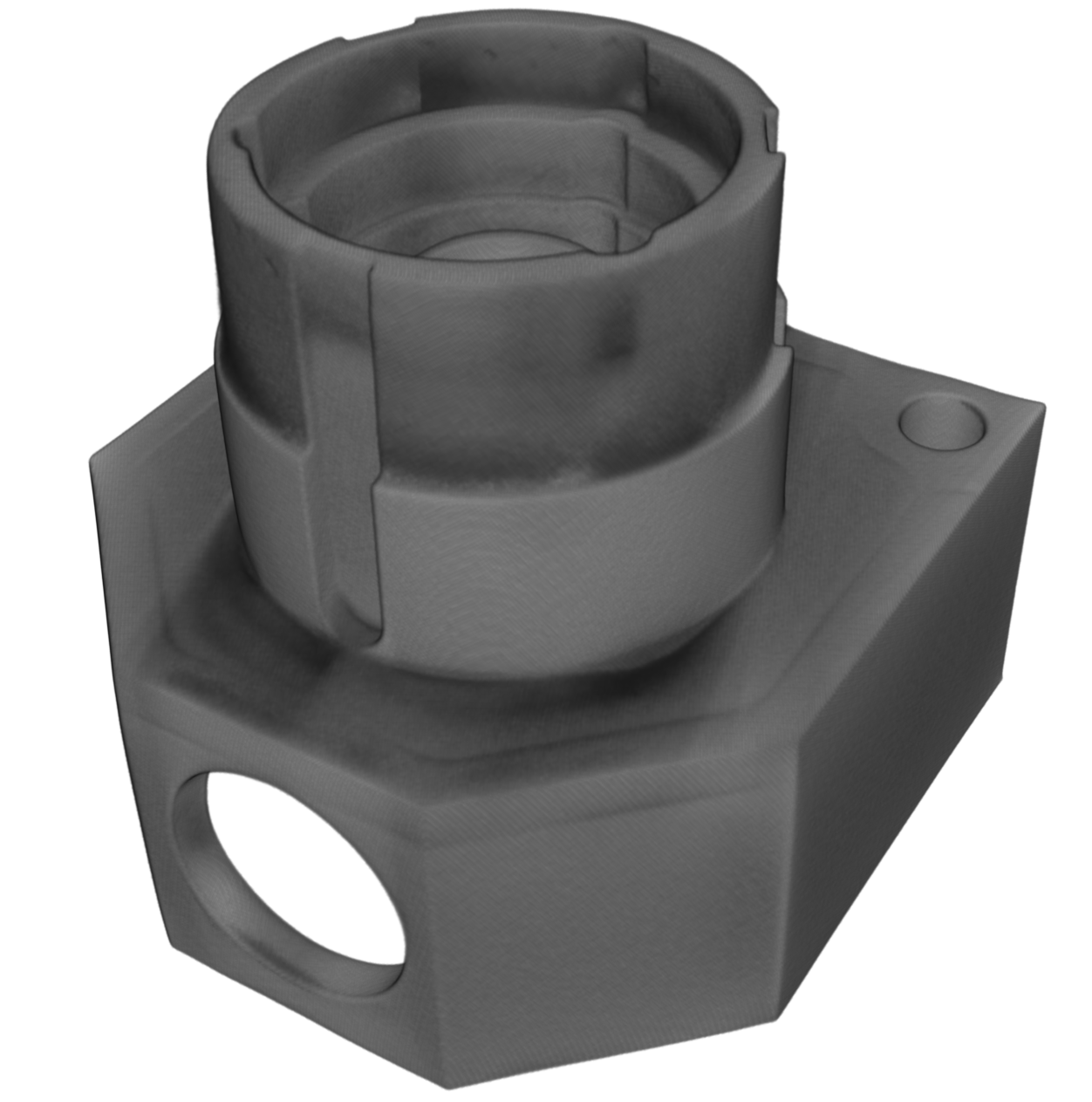}\label{fig:tpa}} \hfil
 \caption{Datasets for which ground truth data is available.~\protect\subref{fig:piston} shows
          a Mahle motor piston,~\protect\subref{fig:bigpiston} incroporates this piston
          in with a "Leberkassemmel" data.~\protect\subref{fig:tasterw} shows a calibration
          dataset, while~\protect\subref{fig:ring} and~\protect\subref{fig:rum} show individual
          parts of a small airplane (courtesy of \emph{Deutsches Museum})
          where the components under investigation are highlighted in white,
          and~\protect\subref{fig:tpa} again is a calibration dataset.%
 }
 \label{fig:quantitativedatasets}
\end{figure*}

\subsubsection*{Influence of multiple resolution levels}
Next, we analyze the influence of multiple resolution levels on
quantitative segmentation results. 
For this purpose, we apply the same kind of evaluation over multiple
resolution levels, i.e., we use the same seed voxels
on all available scales, train and apply the model and again use the
postprocessing specified in Table~\ref{tbl:experimentsetup}. The
results are depicted in Figure~\ref{fig:mrevaluation}.  
Depending on the dataset, the IoU stays approximately constant over
multiple scales which shows the robustness of the approach
for well-suited datasets. On the other hand, this measure deteriorates
a little bit when we aim to segment objects that
have a curved geometry. This observation is easily explained as
especially curvature gets lost when the 
resolution gets  coarser, thus the according regions are not
considered as candidates on a coarse level and hence are not 
segmented on finer ones. We observed that behavior on datasets
including the "Ford Fiesta" and the "Tasterwald" scan. 
There, our goal was to segment the springs and the small orbs,
respectively. In both cases, the desired objects 
have considerable curvature which degrades on a coarser scale.
Recall that the IoU evaluation metric is sensitive to individual
misclassified voxels, which explains the smaller values when small
voxel regions are not segmented using multiple resolutions.
The same trend is reflected by the $F_1$ score evaluation although the
obtained values are still high.
Note that for all four different settings the same postprocessing was
applied  per volume. In practice, however, 
the postprocessing should be interactively adapted to the individual
outcomes for best possible results. This, however, would reduce the
comparability of the result volumes.
We further measured the execution time speedups obtained when using
multiple resolution levels, where we distinguished between the overall
speedup including the training process,
cf. Figure~\ref{fig:evalmrtrain}, and the speedup  
on the evaluation part only, cf. Figure~\ref{fig:evalmrwotrain}. Since in the
multiresolution case we need to train multiple models, we obtain an
increase in the runtime while, at the same time, the real benefit of
multiscale algorithms is obtained in the evaluation part.
If datasets are well-suited for multiresolution processing as
exemplary explained for the Ford Fiesta earlier, 
Algorithm~\ref{alg:mrseg} really achieves considerable speedups of up
to $20\times$ over the single resolution execution. 
On several scans, e.g., "TPA", where nearly the entire volume consists
only of the object that shall be segmented the runtime drops to about
the same as in the single resolution case.
But in such cases the
execution might actually get slower due to the same region but more
voxels need to be processed on each scale.
In conclusion, the localized multiresolution processing enables and
accelerates the segmentation of volumes without size restrictions and
at the cost of only minor performance decreases, depending on the
dataset.

\begin{figure*}[!t]
 \centering
 \subfloat[IoU]{%
  \includegraphics[trim={1cm 1.5cm 2cm 2.8cm}, clip, width=.5\textwidth]{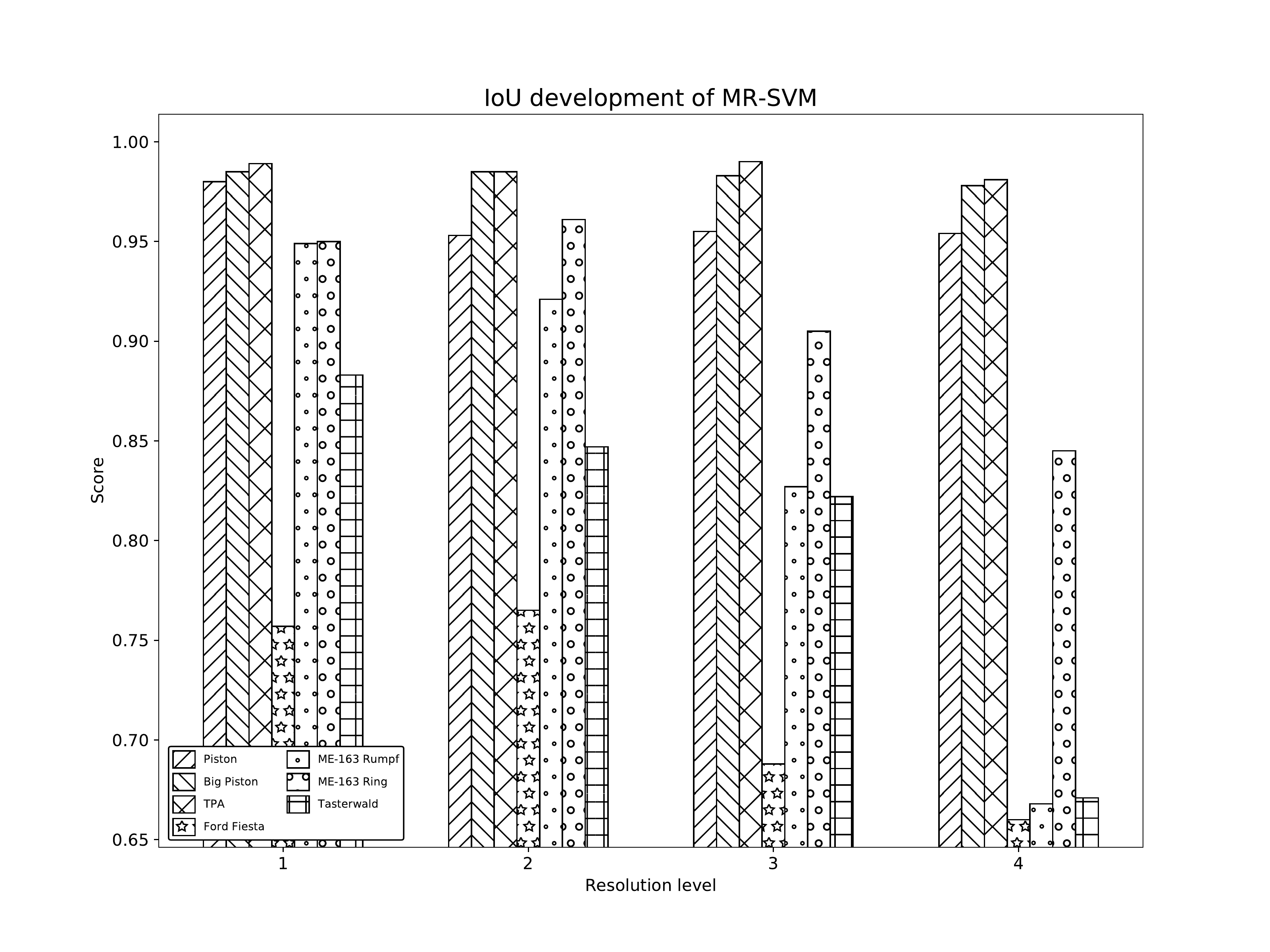}\label{fig:evaliou}}\hfil 
 \subfloat[$F_1$ score]{%
  \includegraphics[trim={1cm 1.5cm 2cm 2.8cm}, clip, width=.5\textwidth]{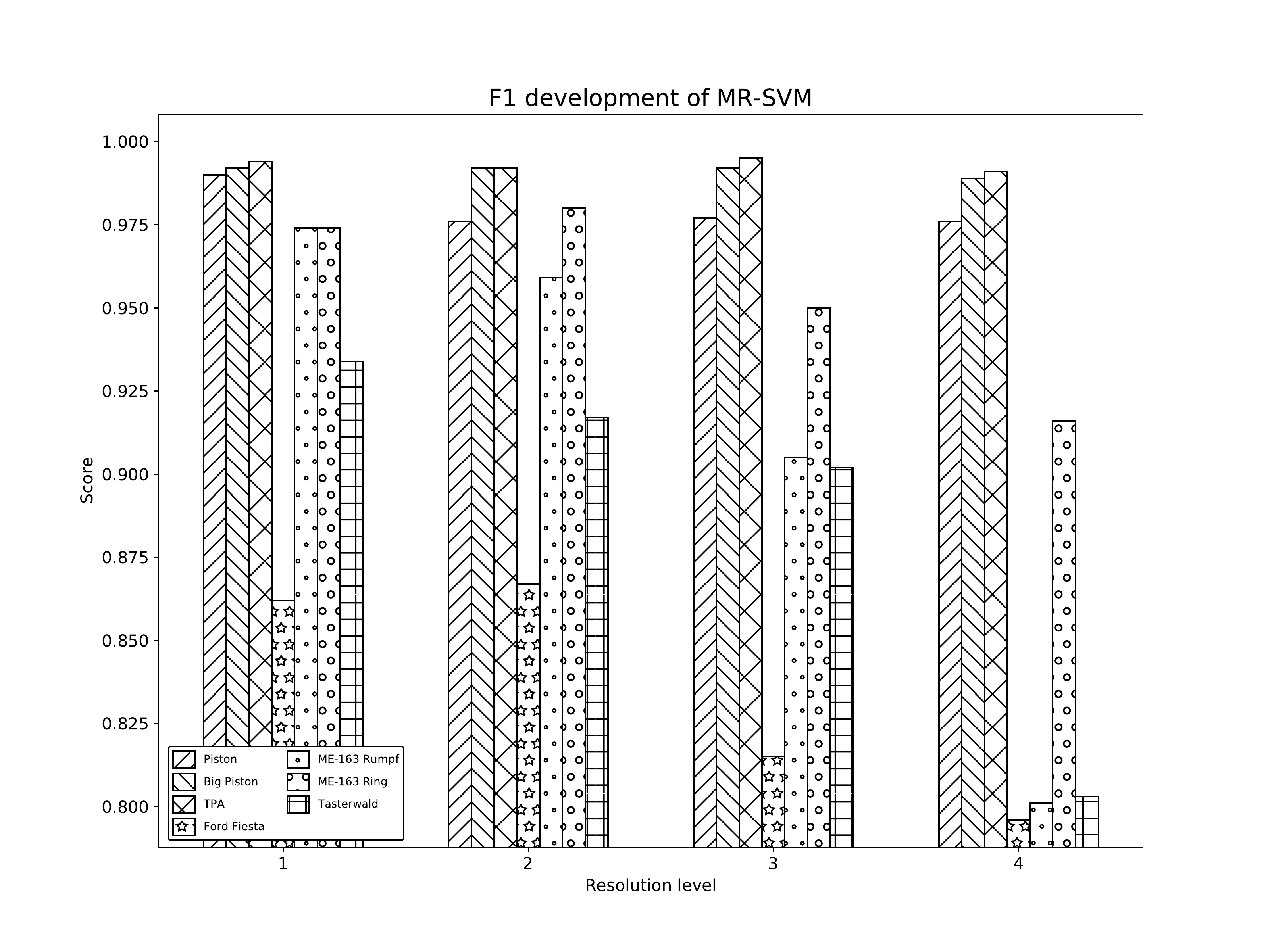}\label{fig:evalf1}}\hfil
 \subfloat[Speedups without training]{%
  \includegraphics[trim={0 1cm 0 1.1cm}, clip, width=.5\textwidth]{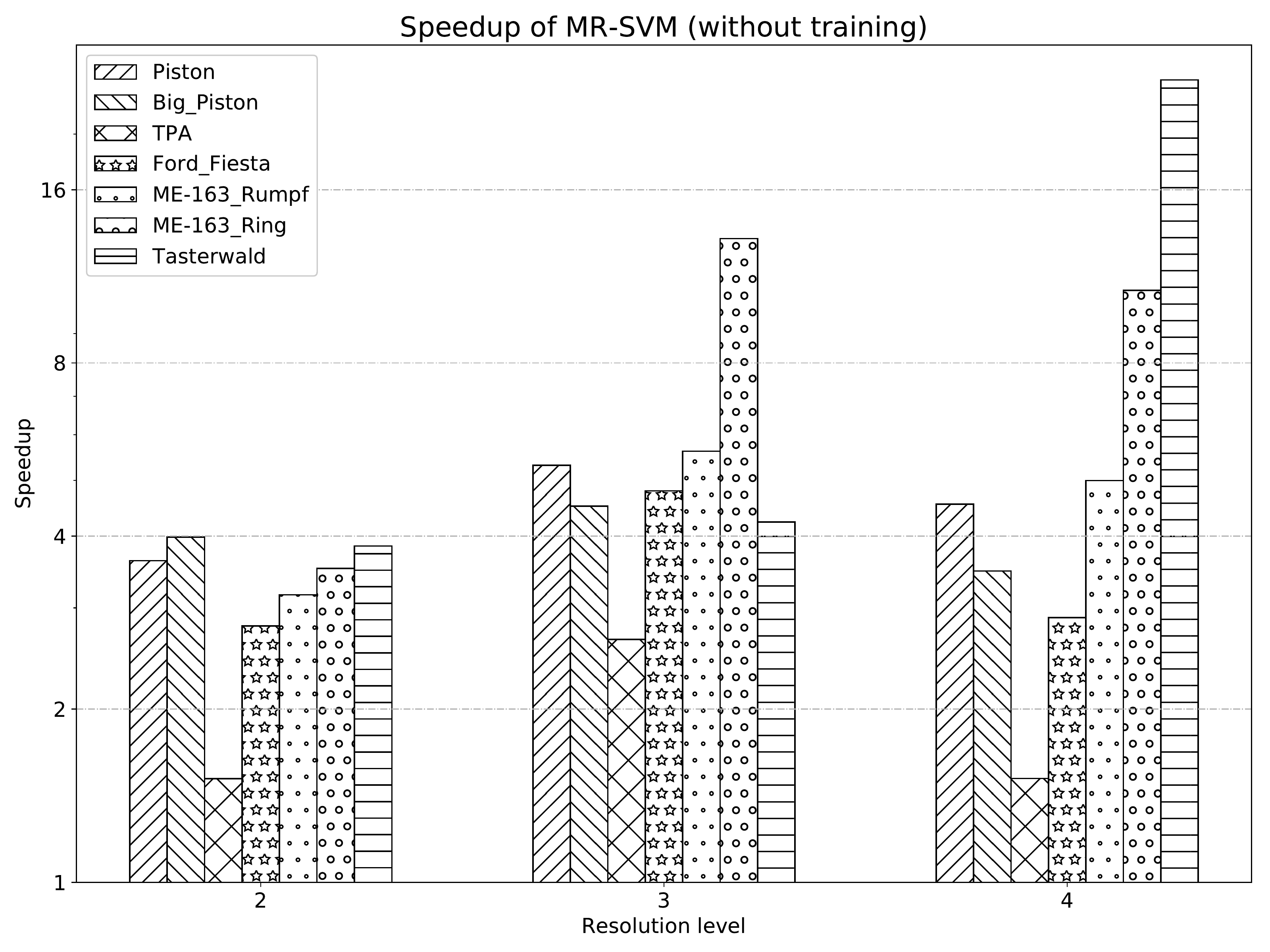}\label{fig:evalmrwotrain}} \hfil 
 \subfloat[Speedups with training]{%
  \includegraphics[trim={0 1cm 0 1.1cm}, clip, width=.5\textwidth]{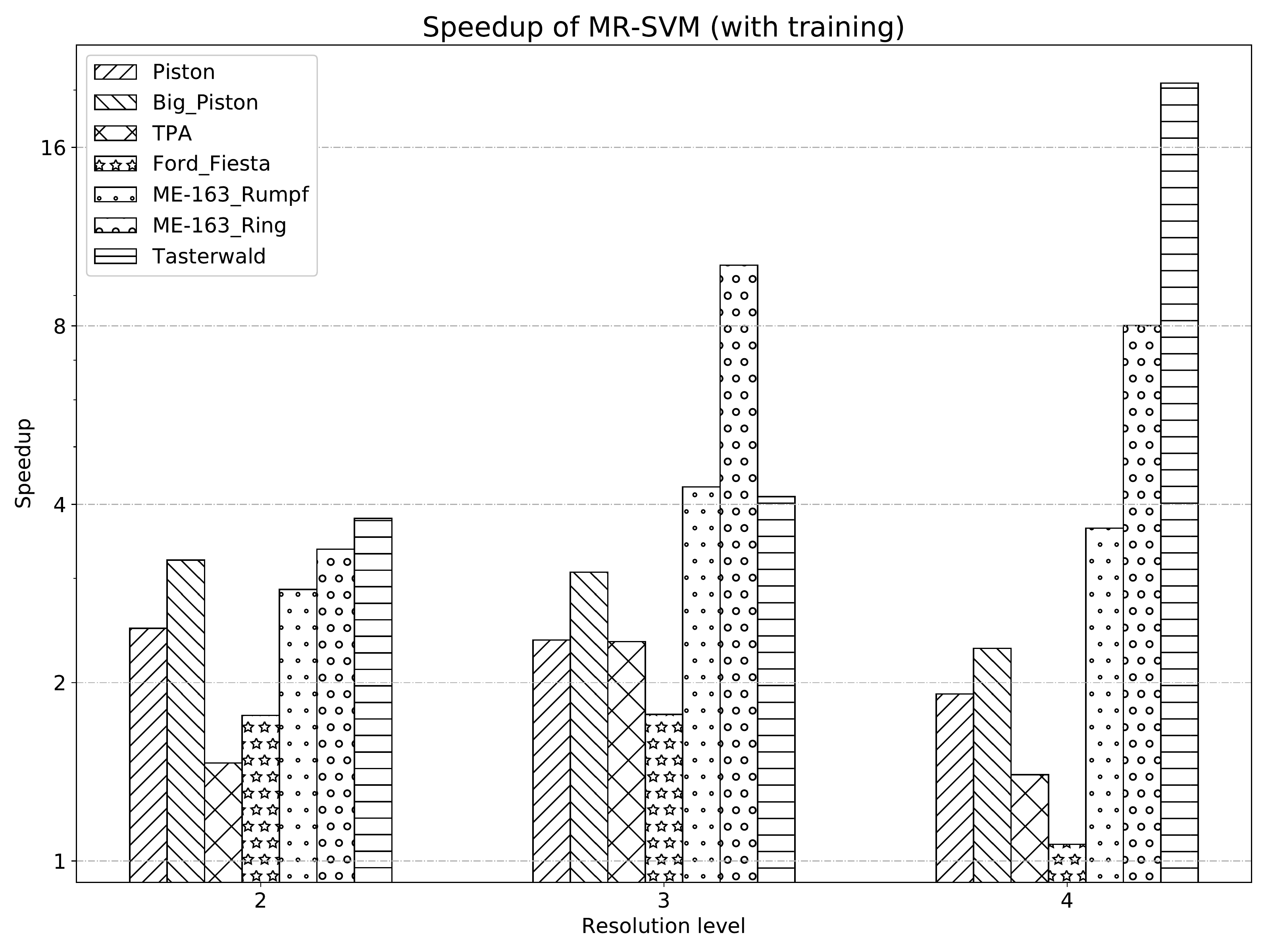}\label{fig:evalmrtrain}} \hfil 
 \caption{Quantitative evaluation over multiple scales.~\protect\subref{fig:evaliou} shows the evaluation
          with respect to IoU while~\protect\subref{fig:evalf1} evaluates w.r.t. the $F_1$ score.
          Contrary,~\protect\subref{fig:evalmrwotrain} and~\protect\subref{fig:evalmrtrain} show the
          speedups for the same datasets when using up to four resolution levels.%
 }
 \label{fig:mrevaluation}
\end{figure*}

\subsection{Qualitative Evaluation}
In order to demonstrate the flexibility of our method, we now apply it
to several datasets which  are large and also unique, i.e., scans with
no similar scans available. Such volumes
cannot easily be processed by methods like neural networks.
Renderings of the datasets and the segmentation results are depicted in
Figure~\ref{fig:qualitative}, the parameter setting we used is
summarized in Table~\ref{tbl:qualsetup}. 
Again, we applied problem-specific postprocessing in form of speckle removal
and connected components labeling.

\begin{table}[!h]
 \renewcommand{\arraystretch}{1.3}
 \caption{Qualitative evaluation setup.}
 \label{tbl:qualsetup}
 \centering
 \begin{tabular}{|l||l|c|c|}
  \hline 
  \multirow{2}{*}{\textbf{Scan}} & \multirow{2}{*}{\textbf{Features}} & \multirow{2}{*}{$K_1$} & \textbf{\#Seeds} \\
                &                   &                &  \textbf{(pos / neg)}     \\ \hline
  Ford Fiesta  & Grayscale + Inertia            & 3 &  65 /  81 \\ \hline
  Honda Accord & Grayscale + Inertia + HOG      & 5 & 110 / 142 \\ \hline
  Wheat plant  & Grayscale + Inertia            & 7 &  79 /  39 \\ \hline
  Mummy        & Grayscale + Inertia + Line fit & 7 & 101 / 128 \\ \hline
 \end{tabular}
\end{table}

The first example is a CT scan of a crashed Ford Fiesta with the task
to extract the four springs, where the front pair was actually damaged
and deformed during the crash. We interactively selected some (65)
seed voxels in the 
spring marked red in the rendering and labeled them as positive, i.e.,
told the system that we want to extract this
component. Moreover, we marked some (81) voxels outside that spring as
negative. Next, we applied  
Algorithm~\ref{alg:segalg} with postprocessing in form of speckle
removal and a connected components 
analysis, see previous section. The result shows the four extracted
springs which were segmented using only the information from the
marked one, encoded in 146 voxels in total.
In another crash car, a Honda Accord, the goal was to segment a
component of the frame that had been damaged during the crash. The
challenge in this case is that this component has a very complex
\emph{global} geometry, locally extracted regions, however, can be
successfully identified with each other and distinguished from regions
that were extracted from unrelated parts of the volume. The result is
overlaid in white and shows how even that complex object can be successfully
segmented interactively. 

Switching to an application in biology/agricultural technology, we
applied our method to a scan of a wheat plant which is planted 
in a pot filled with earth. This scan is difficult to segment due to
low contrast, many connecting components (earth, roots, small rocks)
and the complex geometry on the roots. Nevertheless, the interactively
obtained segmentation result was satisfactory and even able to detect
many very thin roots efficiently.

Finally, in the context of cultural heritage, we consider a scan of a
Peruvian mummy, data courtesy of the Lindenmuseum in Stuttgart,
which is arguably the highest resolution scan ever made of a mummy,
having a voxel edge length of 95 micrometers and being roughly 987
Gigabytes in size. Our focus here lied in
segmenting the ropes from the scan (downsampled by a factor of 4 in
each direction) which hold the bundle together. Again, we marked rope
voxels and labeled them as positive while marking other voxels as
negative. 
Since we have a priori knowledge about the structure we aim to
segment, we made use of it by realizing that it is beneficial
to explicitly consider the line fit feature designed to detect linear
structures, c.f. Table~\ref{tbl:qualsetup}. 
The results are clearly visible ropes and their connections, which we
overlaid over the original data.

\begin{figure*}[!t]
 \centering
 \subfloat[Original]{ \includegraphics[width=.24\textwidth]{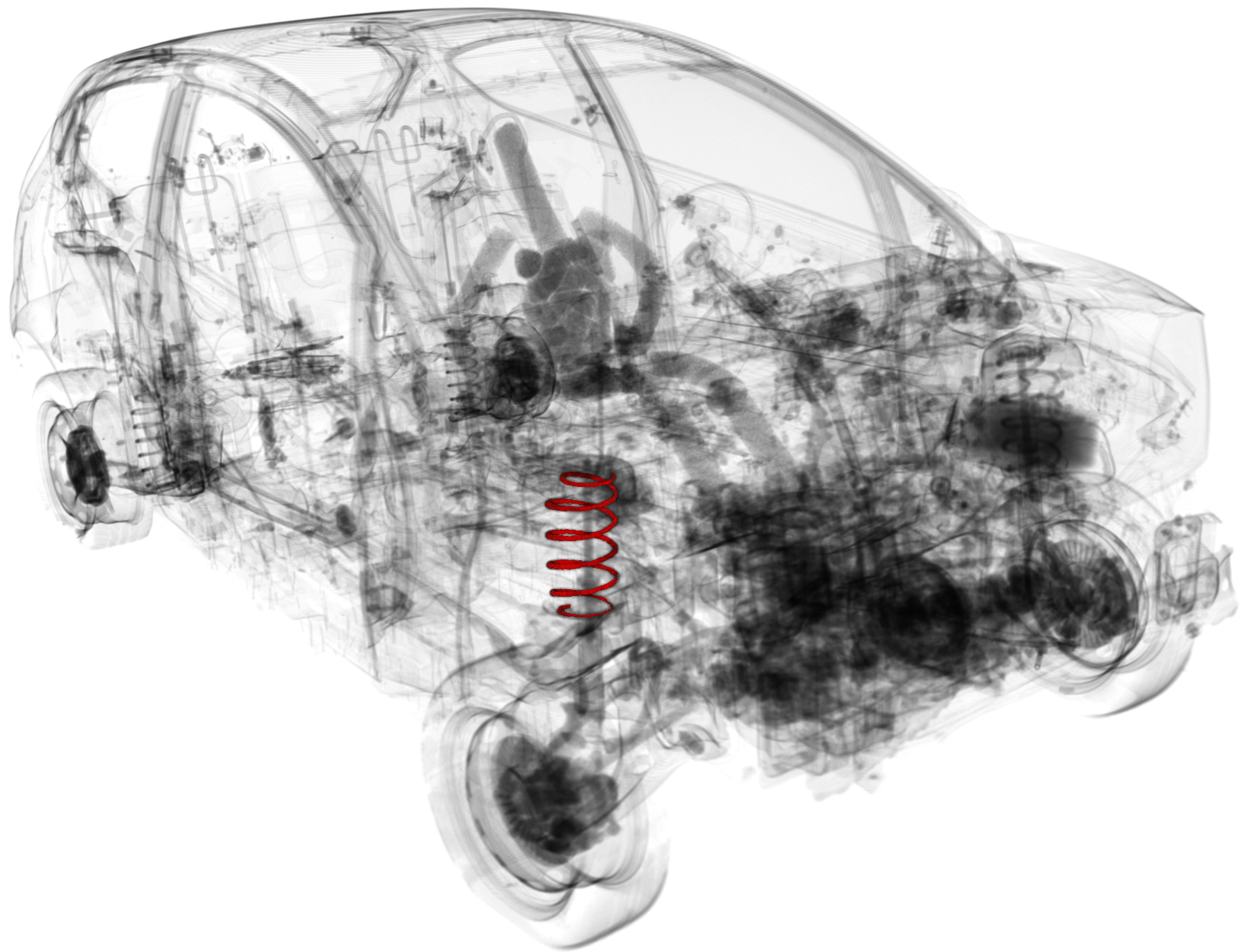}\label{fig:fiesta}} \hfil
 \subfloat[Segmented]{\includegraphics[width=.24\textwidth]{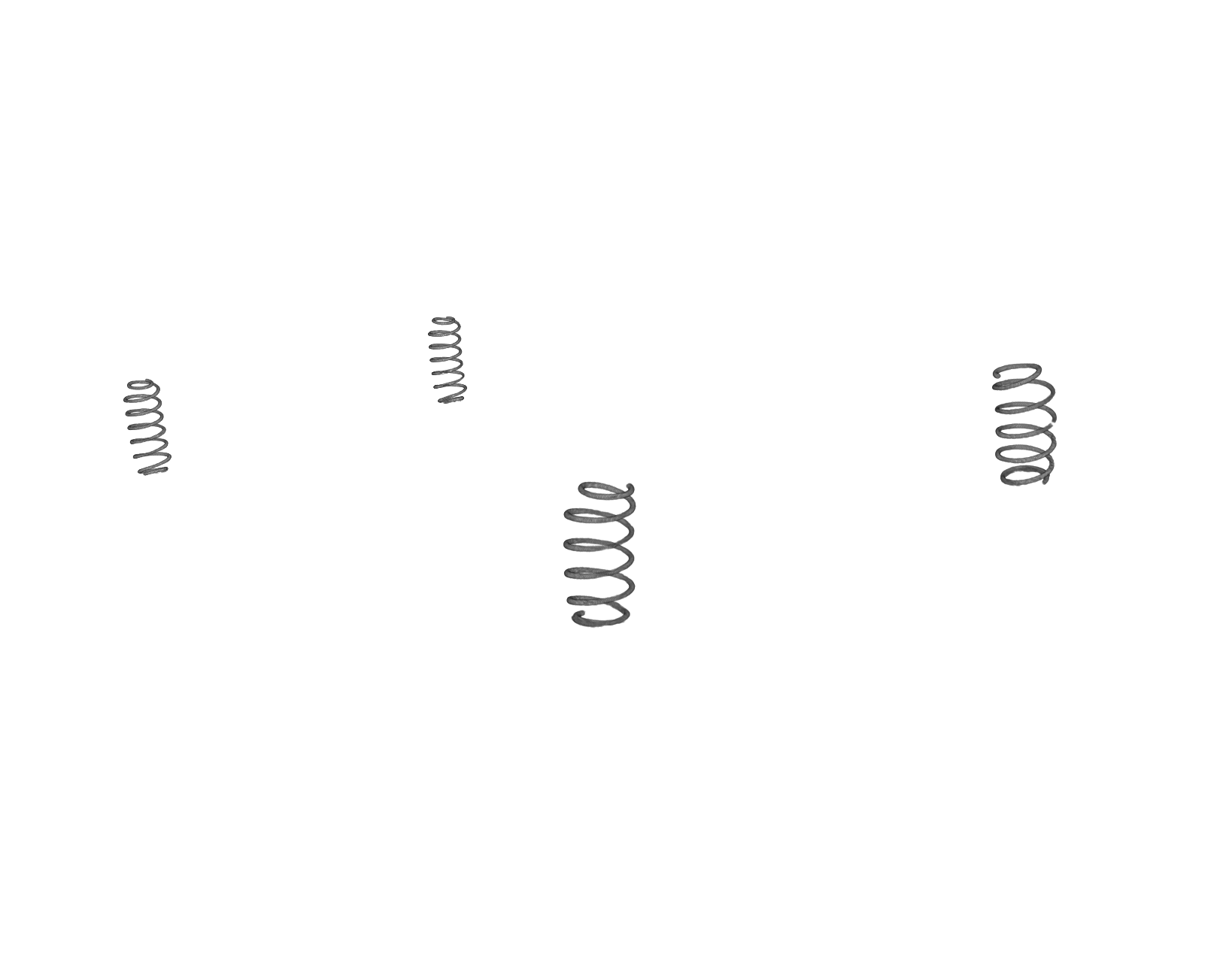}}                          \hfil
 \subfloat[Original]{ \includegraphics[width=.24\textwidth]{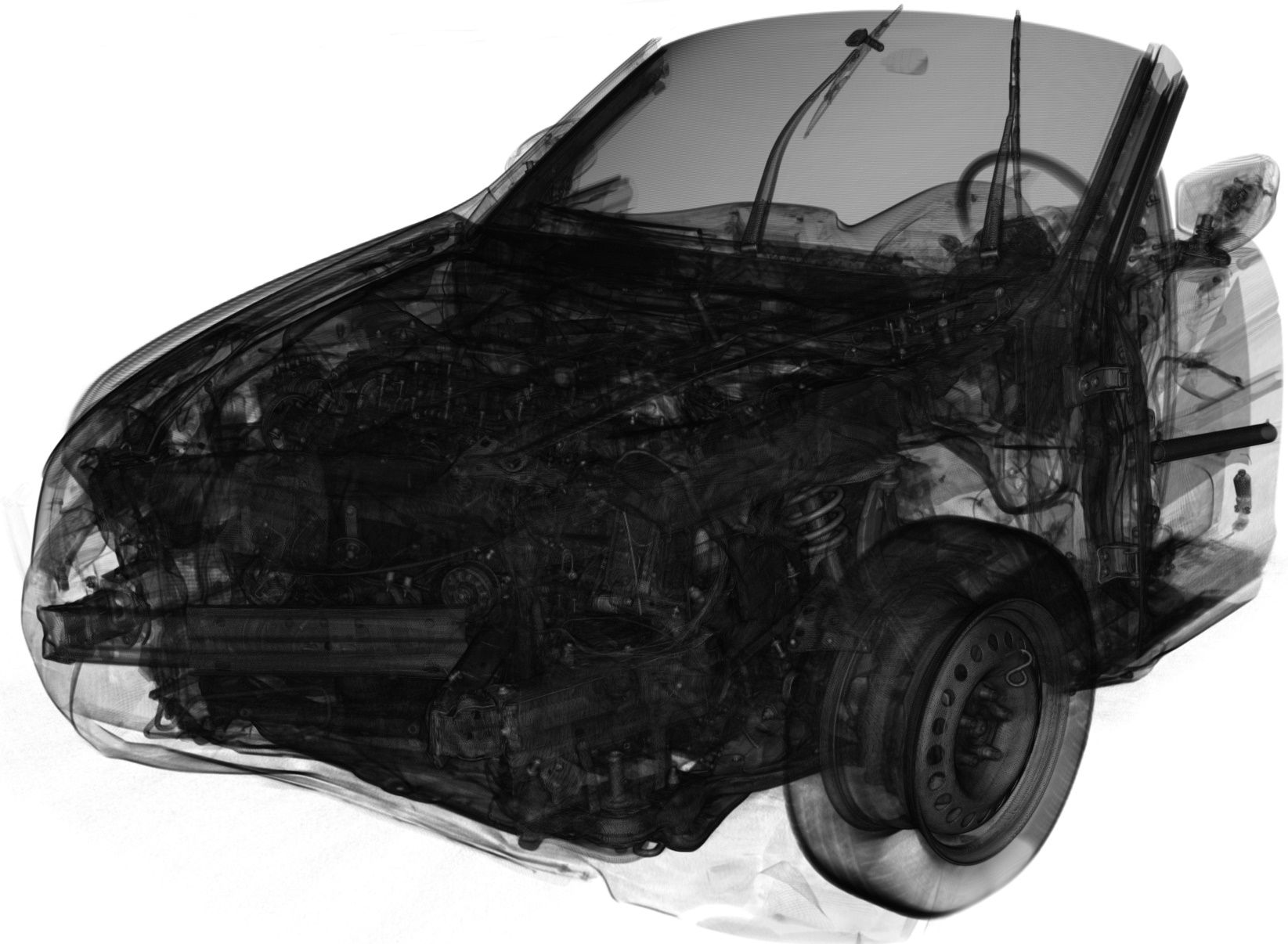}}                      \hfil
 \subfloat[Segmented]{\includegraphics[width=.24\textwidth]{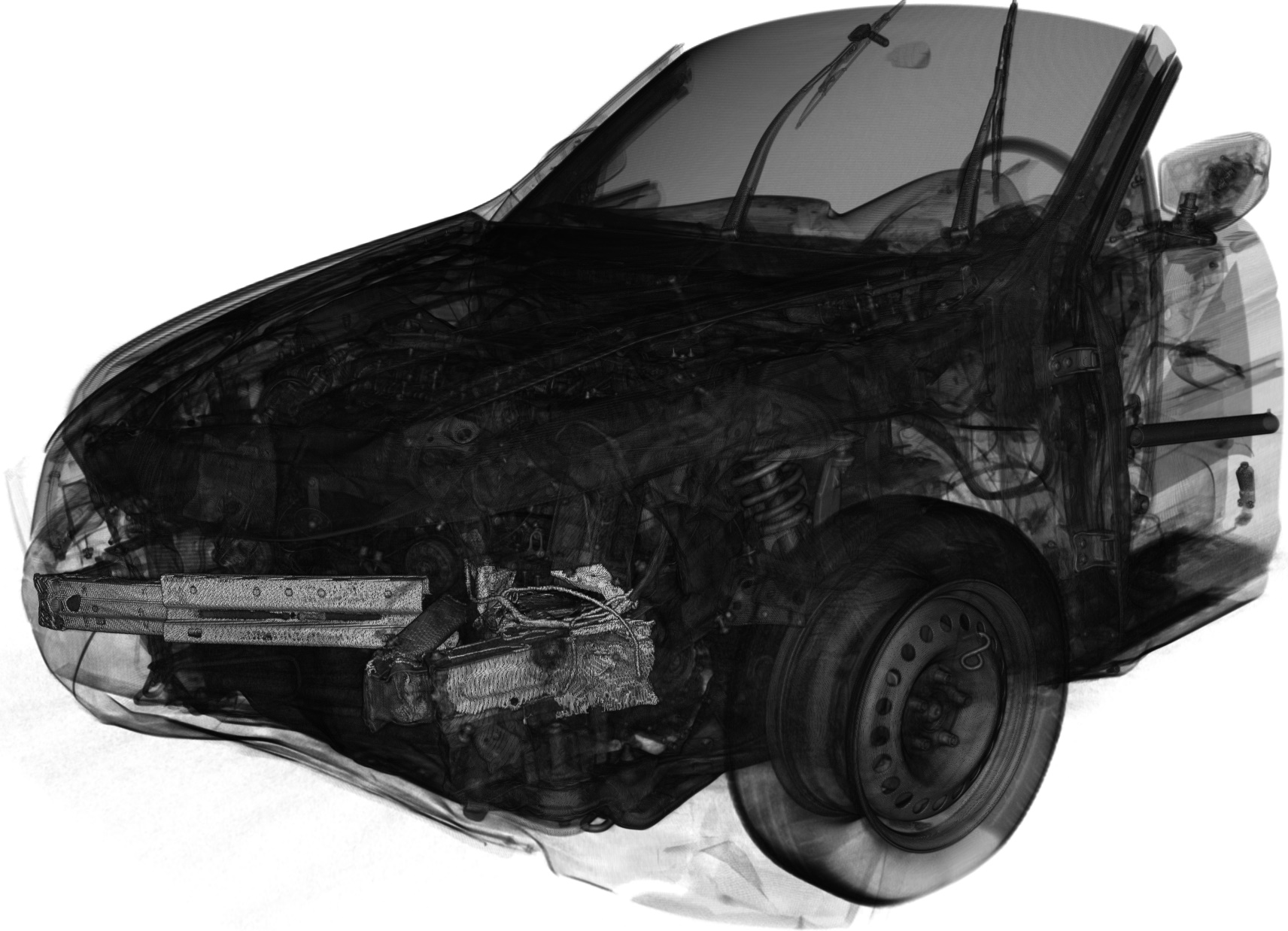}}               \hfil
 \subfloat[Original]{ \includegraphics[width=.21\textwidth]{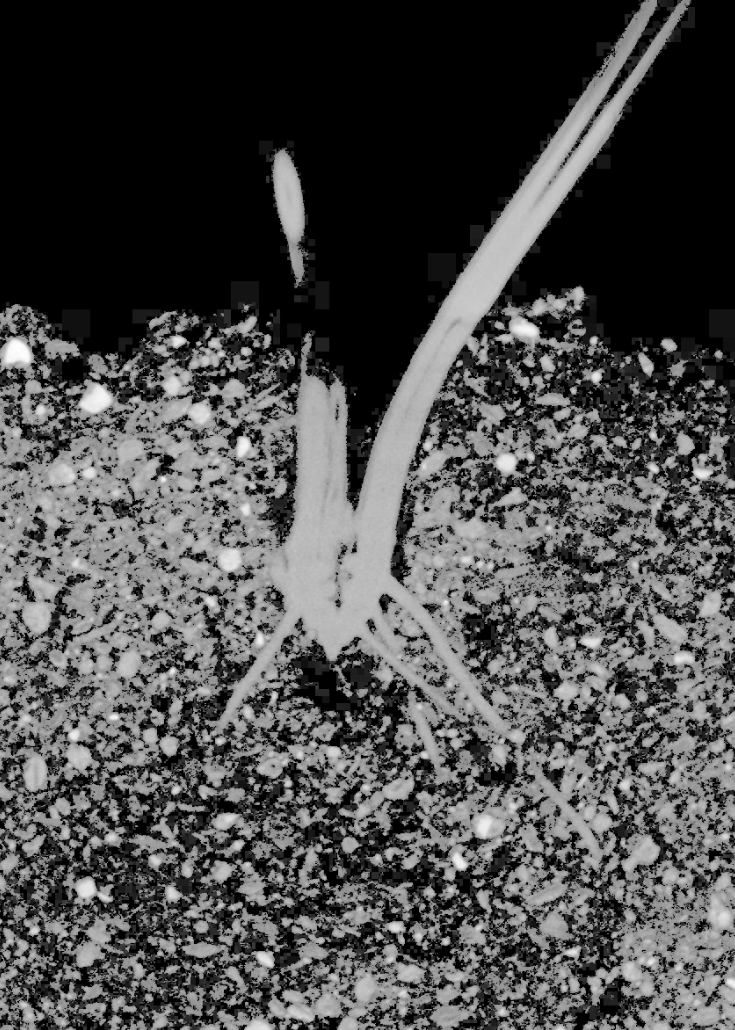}}                       \hfil
 \subfloat[Segmented]{\includegraphics[width=.21\textwidth]{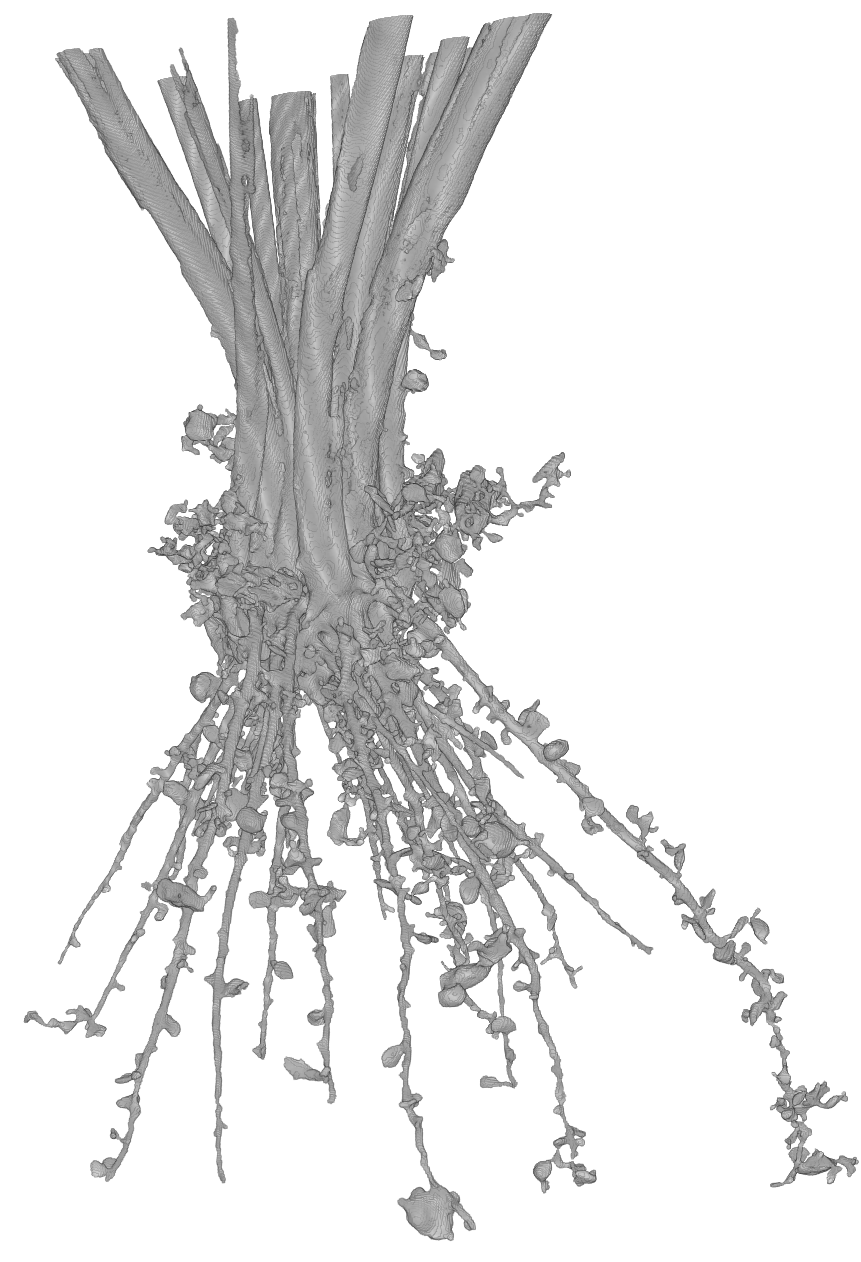}}                   \hfil
 \subfloat[Original]{ \includegraphics[width=.19\textwidth]{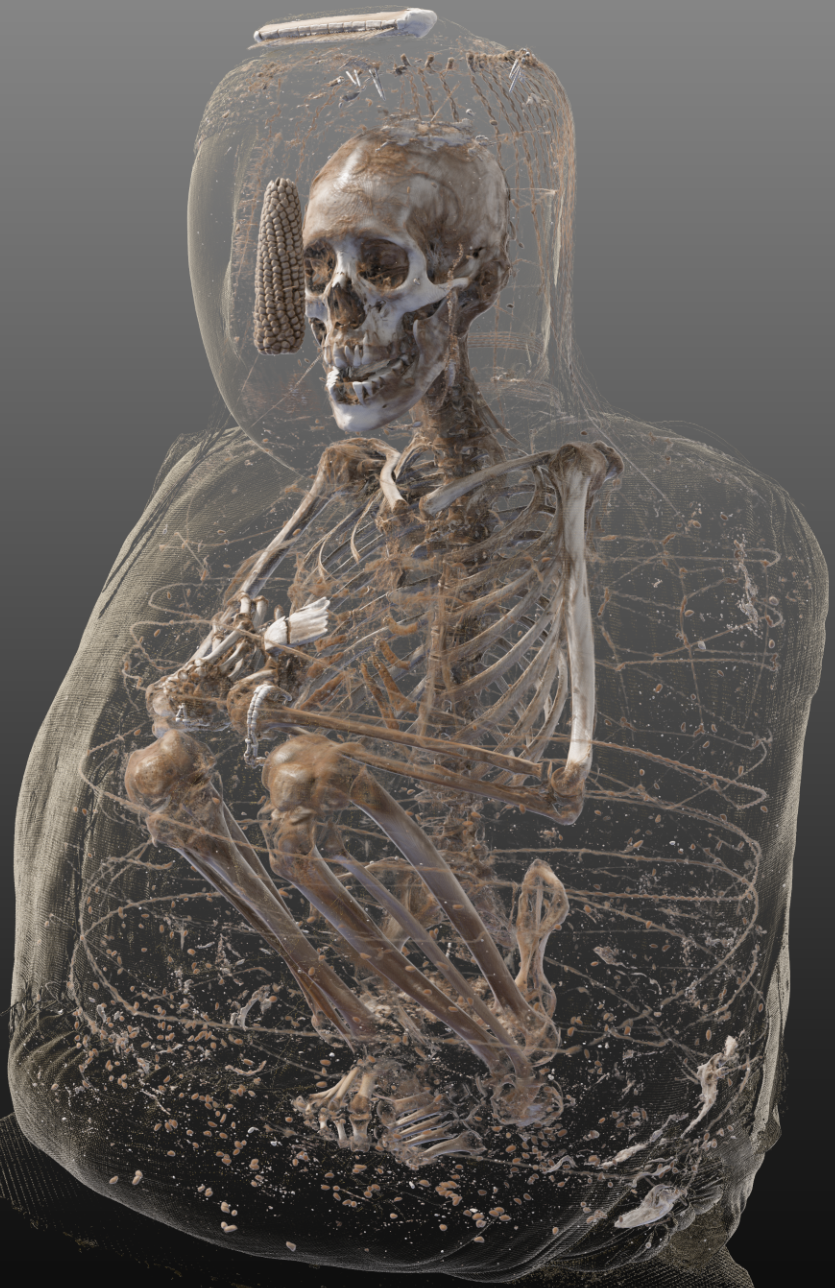}\label{fig:mevismummy}}   \hfil
 \subfloat[Segmented]{\includegraphics[width=.21\textwidth]{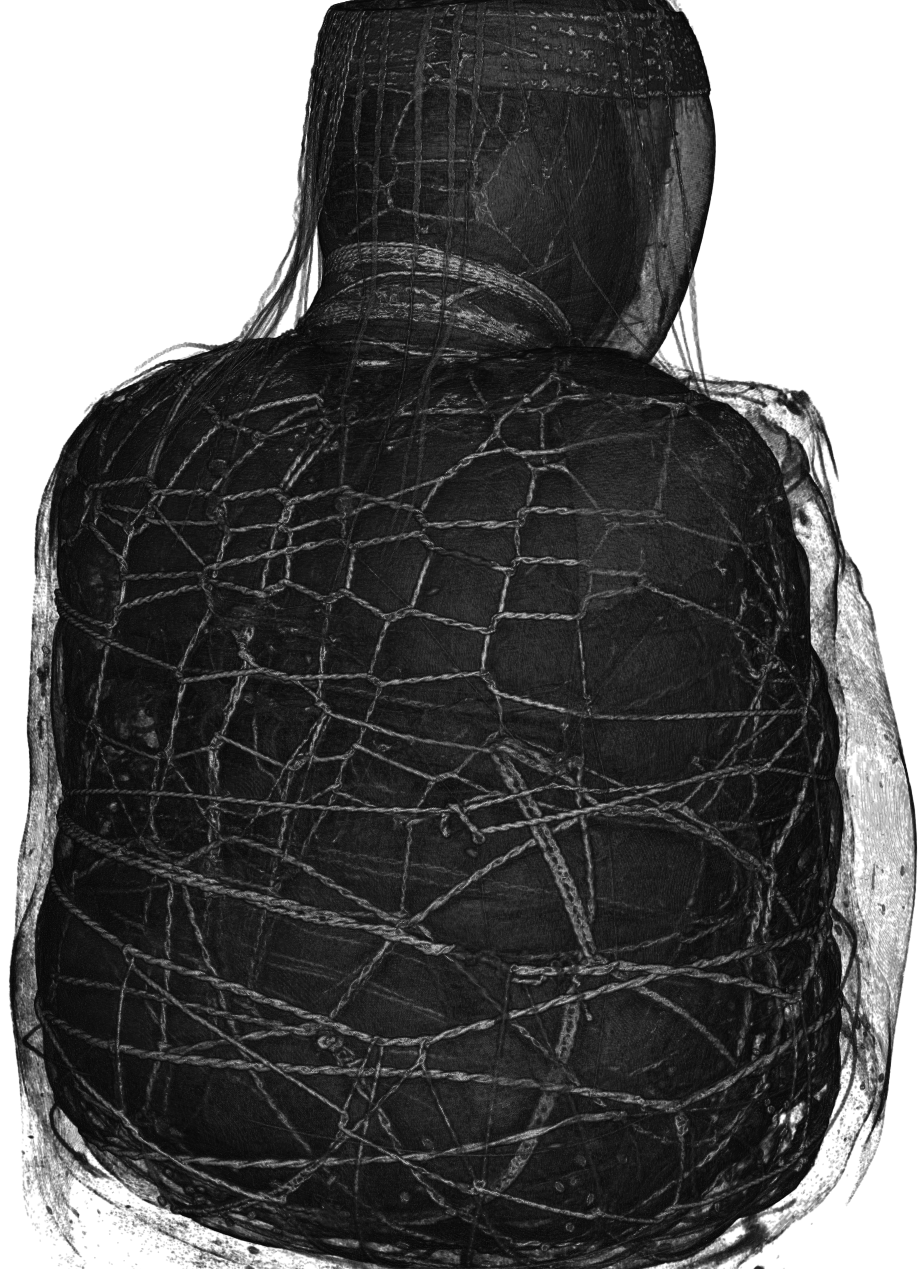}}          \hfil
 \caption{Qualitative segmentation results, each showing first a rendering of the original dataset and
 the segmentation result. The rendering in~\protect\subref{fig:mevismummy} was provided by Fraunhofer MEVIS.
 Contrast was enhanced where necessary. In case of the mummy dataset and the
 second crash car, the segmented components were overlaid over the original voxel dataset.
 The other results are not overlaid for better visibility.}
 \label{fig:qualitative}
\end{figure*}

\section{Conclusion}\label{sec:conclusion}
In industrial computed tomography there is a special interest in the
task of segmentation, i.e., 
partitioning a large three-dimensional voxel 
dataset into several disjoint components with the goal of extracting
information about the object under consideration.
In this task, one faces the challenge that most modern methods require a lot of
annotated training data samples which are rarely available, especially
in unique scans.

We introduced a novel interactive segmentation procedure which
combines active learning, geometric features and local
processing. Besides giving a user a possibility at all to define
\emph{what} should be selected from the volume, the active learning
part gives a flexible way of incorporating domain knowledge ad hoc into the 
algorithm and thus allows us to use the very same algorithm for a big
variety of tasks.
The very general approach of comparing local regions to geometric
primitives mimics the human perception of objects and enables the
application of the proposed scheme on almost any type of scan without
the need to design a separate segmentation method for 
each component to be extracted, as is common practice, e.g., in the
majority of clinical applications.
Finally, the purely local processing not only simplifies computations
but also renders our algorithm applicable to perform three-dimensional
image processing on volumes which may be larger than main memory. 
This marks our procedure as one of the very first methods which were
explicitly designed to be used in the context of large tomography scans
while still maintaining flexibility, since we can use it to segment
many different components. 
On top of that we enhanced the method by a multiresolution processing
algorithm which incorporates both local processing 
and our geometric active learning approach and enables a faster
interactive processing of volumes without 
size restrictions.
In addition, the algorithm works on \emph{voxel datasets}, so we are
not limited to computed tomography but our algorithm can be applied to
any three-dimensional voxel data, also from different domains, like MRI data.

\appendices

\ifCLASSOPTIONcompsoc
  \section*{Acknowledgments}
\else
  \section*{Acknowledgment}
\fi

This work was
supported by the project 
"Digitalisierung, Verarbeitung und Analyse kultureller und industrieller
Objekte: Wertsch{\"o}pfung aus großen Daten und Datenmengen - Big Picture" that has been promoted and funded by
the Bavarian State Government under Grant No. AZ.: 43-6623/138/2 since 2018 until 2021.

\bibliographystyle{IEEEtran}
\bibliography{\jobname}

\begin{thebibliography}{10}
\providecommand{\url}[1]{#1}
\csname url@samestyle\endcsname
\providecommand{\newblock}{\relax}
\providecommand{\bibinfo}[2]{#2}
\providecommand{\BIBentrySTDinterwordspacing}{\spaceskip=0pt\relax}
\providecommand{\BIBentryALTinterwordstretchfactor}{4}
\providecommand{\BIBentryALTinterwordspacing}{\spaceskip=\fontdimen2\font plus
\BIBentryALTinterwordstretchfactor\fontdimen3\font minus
  \fontdimen4\font\relax}
\providecommand{\BIBforeignlanguage}[2]{{%
\expandafter\ifx\csname l@#1\endcsname\relax
\typeout{** WARNING: IEEEtran.bst: No hyphenation pattern has been}%
\typeout{** loaded for the language `#1'. Using the pattern for}%
\typeout{** the default language instead.}%
\else
\language=\csname l@#1\endcsname
\fi
#2}}
\providecommand{\BIBdecl}{\relax}
\BIBdecl

\bibitem{UNet}
{\"O}.~{\c{C}}i{\c{c}}ek, A.~Abdulkadir, S.~S. Lienkamp, T.~Brox, and
  O.~Ronneberger, ``{3D U-Net: Learning Dense Volumetric Segmentation from
  Sparse Annotation},'' in \emph{Medical Image Computing and Computer-Assisted
  Intervention -- MICCAI 2016}, S.~Ourselin, L.~Joskowicz, M.~R. Sabuncu,
  G.~Unal, and W.~Wells, Eds., 2016, pp. 424--432.

\bibitem{DLAL}
L.~Yang, Y.~Zhang, J.~Chen, S.~Zhang, and D.~Z. Chen, ``{Suggestive Annotation:
  {A} Deep Active Learning Framework for Biomedical Image Segmentation},''
  \emph{CoRR}, vol. abs/1706.04737, 2017.

\bibitem{ViewAL}
Y.~Siddiqui, J.~Valentin, and M.~Nie{\ss}ner, ``{ViewAL: Active Learning with
  Viewpoint Entropy for Semantic Segmentation},'' \emph{CoRR}, vol.
  abs/1911.11789, 2019.

\bibitem{GANAL}
D.~Mahapatra, B.~Bozorgtabar, J.~Thiran, and M.~Reyes, ``{Efficient Active
  Learning for Image Classification and Segmentation using a Sample Selection
  and Conditional Generative Adversarial Network},'' \emph{CoRR}, vol.
  abs/1806.05473, 2018.

\bibitem{SVMAL1}
S.~Tong and D.~Koller, ``{Support Vector Machine Active Learning with
  Applications to Text Classification},'' \emph{J. Mach. Learn. Res.}, vol.~2,
  pp. 45–--66, 2002.

\bibitem{ActiveLearningSurvey}
B.~Settles, ``Active learning literature survey,'' University of
  Wisconsin--Madison, Computer Sciences Technical Report 1648, 2009.

\bibitem{SVMAL2}
J.~Kremer, K.~Steenstrup~Pedersen, and C.~Igel, ``{Active Learning with Support
  Vector Machines},'' \emph{WIREs Data Min. Knowl. Discov.}, vol.~4, no.~4, pp.
  313--326, 2014.

\bibitem{SVMAL3}
L.~Wang, K.~L. Chan, and Z.~Zhang, ``{Bootstrapping SVM active learning by
  incorporating unlabelled images for image retrieval},'' in \emph{IEEE
  Computer Society Conference on Computer Vision and Pattern Recognition},
  2003, pp. 629--634.

\bibitem{SVMAL4}
G.~Matasci, D.~Tuia, and M.~Kanevski, ``Svm-based boosting of active learning
  strategies for efficient domain adaptation,'' \emph{IEEE Journal of Selected
  Topics in Applied Earth Observations and Remote Sensing}, vol.~5, no.~5, pp.
  1335--1343, 2012.

\bibitem{SVMAL5}
H.~Veeraraghavan and J.~V. Miller, ``{Active Learning Guided Interactions for
  Consistent Image Segmentation with Reduced User Interactions},'' in
  \emph{IEEE International Symposium on Biomedical Imaging: From Nano to
  Macro}, 2011, pp. 1645--1648.

\bibitem{SVMInteractive2D}
B.~Mathieu, A.~Crouzil, and J.~Puel, ``{Interactive multiclass segmentation
  using superpixel classification},'' \emph{CoRR}, vol. abs/1510.03199, 2015.

\bibitem{SVMInteractive3D}
X.~Zhang, J.~Tian, D.~Xiang, X.~Li, and K.~Deng, ``{Interactive liver tumor
  segmentation from ct scans using support vector classification with
  watershed},'' in \emph{Annual International Conference of the IEEE
  Engineering in Medicine and Biology Society}, 2011, pp. 6005--6008.

\bibitem{GeomAL}
K.~Konyushkova, R.~Sznitman, and P.~Fua, ``{Geometry in Active Learning for
  Binary and Multi-class Image Segmentation},'' \emph{Comput. Vis. Image
  Underst.}, vol. 182, pp. 1--16, 2019.

\bibitem{FeatureSelectionPCA}
F.~{Song}, Z.~{Guo}, and D.~{Mei}, ``{F}eature {S}election {U}sing {P}rincipal
  {C}omponent {A}nalysis,'' in \emph{International Conference on System
  Science, Engineering Design and Manufacturing Informatization (ICSEM)},
  vol.~1, 2010, pp. 27--30.

\bibitem{OnePassStatisticalMoments}
P.~Pebay and P.~Pierre, ``{Formulas for Robust, One-Pass Parallel Computation
  of Covariances and Arbitrary-Order Statistical Moments},'' U.S. Department of
  Energy, Office of Scientific and Technical Information, Tech. Rep., 2008.

\bibitem{LBPTOP}
G.~Zhao and M.~Pietik{\"a}inen, ``{D}ynamic {T}exture {R}ecognition {U}sing
  {L}ocal {B}inary {P}atterns with an {A}pplication to {F}acial
  {E}xpressions,'' \emph{IEEE Trans. Pattern Anal. Mach. Intell.}, vol.~29, pp.
  915--28, 2007.

\bibitem{LBPMotiv}
O.~Lahdenoja, J.~Poikonen, and M.~Laiho, ``{T}owards {U}nderstanding the
  {F}ormation of {U}niform {L}ocal {B}inary {P}atterns,'' \emph{ISRN Mach.
  Vis.}, vol. 2013, 2013.

\bibitem{LBPUniform}
------, ``{Towards Understanding the Formation of Uniform Local Binary
  Patterns},'' \emph{ISRN Mach. Vis.}, vol. 2013, 2013.

\bibitem{CurvatureWithoutIso}
H.~Yoshida and J.~N{\"{a}}ppi, ``{Three-Dimensional Computer-Aided Diagnosis
  Scheme for Detection of Colonic Polyps},'' \emph{{IEEE} Trans. Med. Imaging},
  vol.~20, no.~12, pp. 1261--1274, 2001.

\bibitem{OrthogonalDistance}
S.~J. Ahn, \emph{{Least Squares Orthogonal Distance Fitting of Curves and
  Surfaces in Space}}, ser. Lect. Notes Comput. Sci.\hskip 1em plus 0.5em minus
  0.4em\relax Springer, 2004, vol. 3151.

\bibitem{OrientationFeatures}
B.~J{\"a}hne, \emph{{Spatial-Temporal Image Processing, Theory and Scientific
  Applications}}.\hskip 1em plus 0.5em minus 0.4em\relax Springer-Verlag Berlin
  Heidelberg, 1993.

\bibitem{LocalFourierFeatures}
A.~Bhalerao and R.~Wilson, ``{A Fourier Approach to 3D Local Feature Estimation
  from Volume Data},'' \emph{Proceedings of BMVC 2001}, 04 2002.

\bibitem{zscore}
E.~Kreyszig, \emph{{Advanced Engineering Mathematics}}, 10th~ed.\hskip 1em plus
  0.5em minus 0.4em\relax Wiley, 2011.

\bibitem{Billingsley}
P.~Billingsley, \emph{{Probability and Measure}}, 3rd~ed., ser. Series in
  Probability and Mathematical Statistics.\hskip 1em plus 0.5em minus
  0.4em\relax Wiley, 1995.

\bibitem{StatisticsInWassersteinSpace}
{Panaretos, Victor and Zemel, Yoav}, \emph{{An Invitation to Statistics in
  Wasserstein Space}}.\hskip 1em plus 0.5em minus 0.4em\relax Springer, 2020.

\bibitem{VapnikSVNetworks}
C.~Cortes and V.~Vapnik, ``Support-vector networks,'' \emph{Mach. Learn.},
  vol.~20, pp. 273--297, 1995.

\bibitem{LIBSVMPaper}
C.-C. Chang and C.-J. Lin, ``{LIBSVM}: A library for support vector machines,''
  \emph{ACM Trans. Intell. Syst. Technol.}, vol.~2, pp. 1--27, 2011.

\bibitem{InterpretationNu}
------, ``{T}raining $\nu$-{S}upport {V}ector {C}lassifiers: {T}heory and
  {A}lgorithms,'' \emph{Neural Comput.}, vol.~13, no.~9, pp. 2119–--2147,
  2001.

\bibitem{PlattScaling}
J.~C. Platt, ``{Probabilistic Outputs for Support Vector Machines and
  Comparisons to Regularized Likelihood Methods},'' in \emph{Advances in
  Large-Margin Classifiers}.\hskip 1em plus 0.5em minus 0.4em\relax MIT Press,
  1999, pp. 61--74.

\bibitem{SVMAdvantages3}
L.~Auria and R.~A. Moro, ``{Support Vector Machines (SVM) as a Technique for
  Solvency Analysis},'' DIW Berlin, German Institute for Economic Research,
  Discussion Papers of DIW Berlin 811, 2008.

\bibitem{SVMCubic}
L.~Bottou, O.~Chapelle, D.~DeCoste, and J.~Weston, \emph{{Support Vector
  Machine Solvers}}.\hskip 1em plus 0.5em minus 0.4em\relax MIT Press, 2007,
  pp. 1--27.

\bibitem{MR1}
Y.~Wang and M.~Yin, ``Multiresolution and multiscale geometric analysis based
  breast cancer diagnosis using weighted svm,'' in \emph{Proceedings of the
  2015 International Conference on Mechanical Science and Engineering}.\hskip
  1em plus 0.5em minus 0.4em\relax Atlantis Press, 03 2016, pp. 373--378.

\bibitem{MR3}
H.~Akbari and B.~Fei, ``3d ultrasound image segmentation using wavelet support
  vector machines,'' \emph{Med Phys}, vol.~39, no.~6, pp. 2972--2984, 6 2012.

\bibitem{MR2}
M.~A.-M.~M. Salem, ``Multiresolution image segmentation,'' Ph.D. dissertation,
  Humboldt-Universit{\"a}t zu Berlin, Mathematisch-Naturwissenschaftliche
  Fakult{\"a}t II, 2008.

\bibitem{MR4}
C.~C. {Reyes Aldasoro} and A.~{Bhalerao}, ``Volumetric texture segmentation by
  discriminant feature selection and multiresolution classification,''
  \emph{IEEE Transactions on Medical Imaging}, vol.~26, no.~1, pp. 1--14, 2007.

\bibitem{MR5}
A.~R. Fuller, R.~Zawadzki, S.~Choi, D.~F. Wiley, J.~S. Werner, and B.~Hamann,
  ``Segmentation of three-dimensional retinal image data,'' \emph{IEEE
  Transactions on Visualization and Computer Graphics}, vol.~13, 2007.

\bibitem{MR6}
M.~de~Bruijne, B.~van Ginneken, M.~A. Viergever, and W.~J. Niessen,
  ``Interactive segmentation of abdominal aortic aneurysms in cta images,''
  \emph{Med. Image Anal.}, vol.~8, no.~2, pp. 127--138, 2004.

\end{thebibliography}

\begin{IEEEbiography}[{\includegraphics[width=1in]{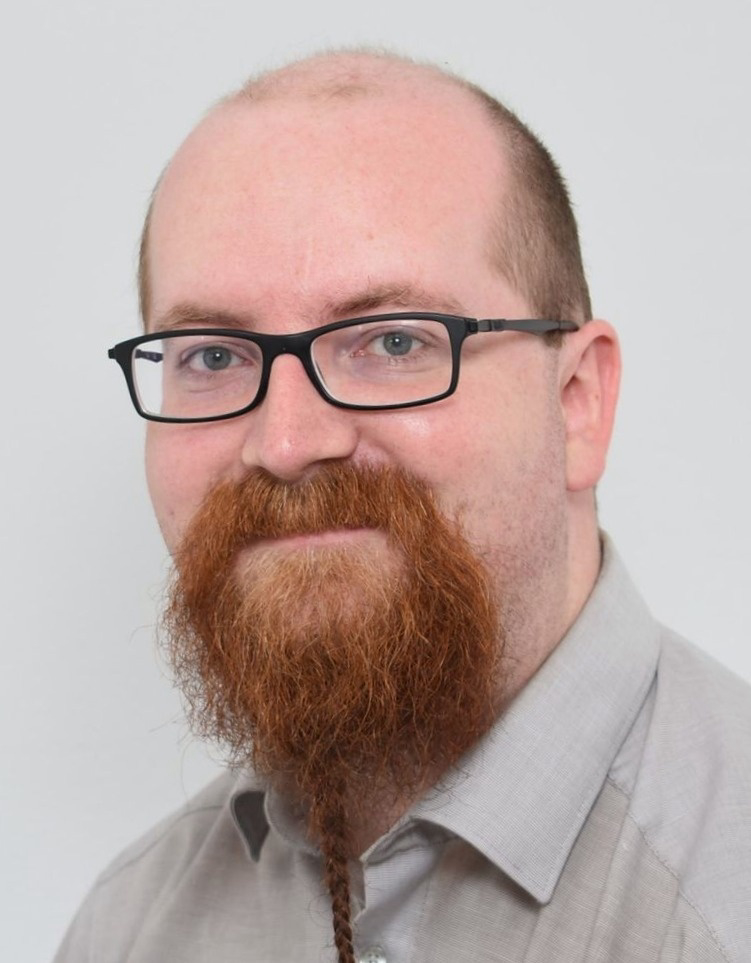}}]{Thomas Lang}
received his PhD degree in computer science from the University of Passau in 2021 for
research in image segmentation. Currently, he is working as a Post-Doc in the
Fraunhofer IIS Research Group on "Knowledge-Based Image Processing".
His research interests include image processing of huge data,
image segmentation and artificial intelligence.
\end{IEEEbiography}

\begin{IEEEbiography}[{\includegraphics[width=1in]{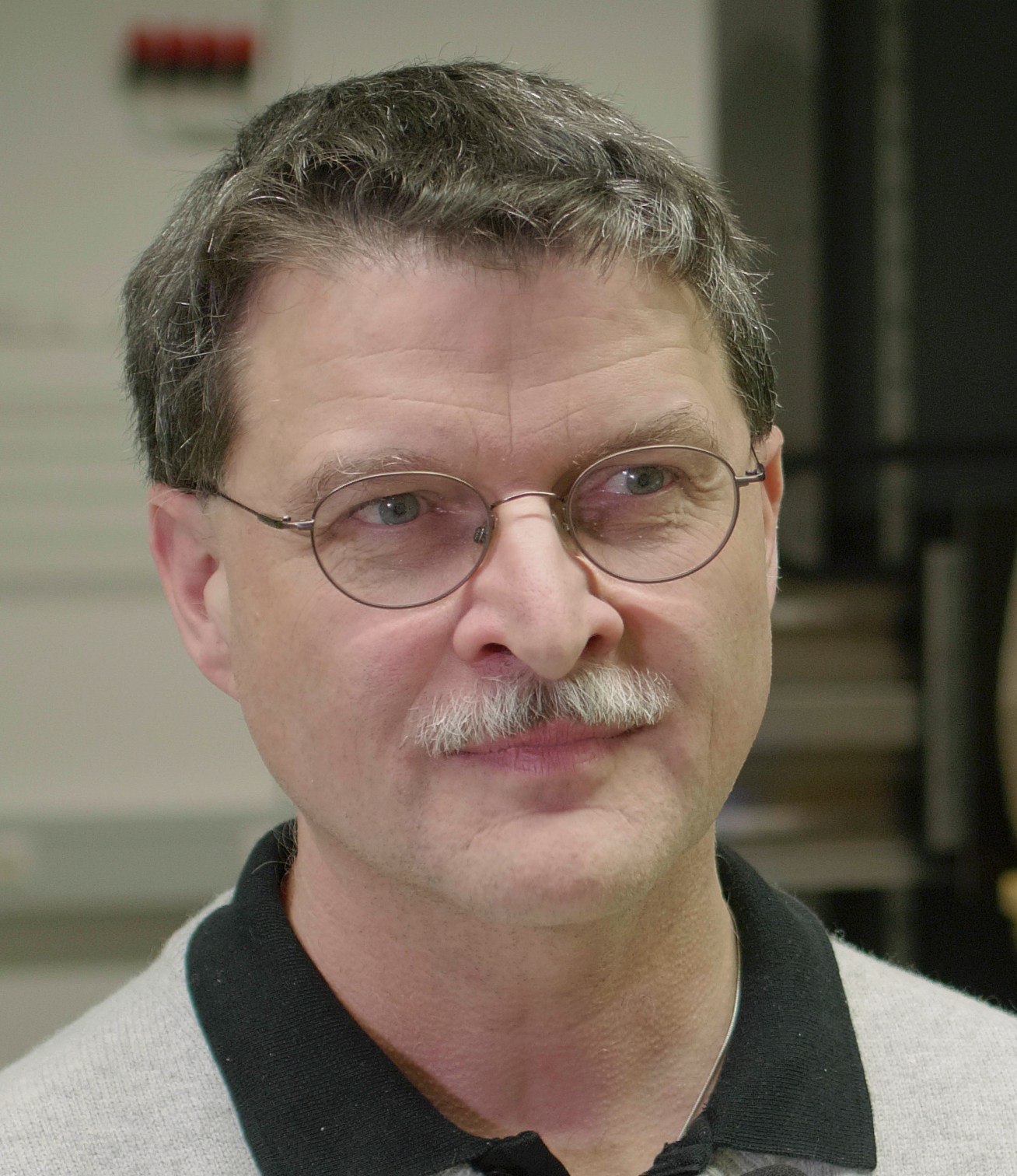}}]{Tomas Sauer}
received his PhD degree in mathematics from the University of
Erlangen-N\"urnberg in 1993 for research in Approximation Theory. From
2000 to 2012, he was professor for Numerical Mathematics/Scientific
Computating at the University of Gie{\ss}en. Since 2012, 
he holds the chair for Mathematical Image Processing at the University
of Passau and is also director of the Applied Research Institute
FORWISS and of the Fraunhofer IIS/EZRT Research Group on
"Knowledge-Based Image Processing". His current research interests 
include, among others, signal and image processing for huge data,
tomography, machine learning, sparse reconstructions, and continued
fractions. 
\end{IEEEbiography}

\end{document}